 \documentclass[final,5p,times,twocolumn]{elsarticle}

\usepackage{graphicx}
\usepackage{amssymb}
\usepackage{array}
\usepackage{subfigure}
\usepackage{epsfig}
\usepackage{comment}
\graphicspath{{./images/}}

\usepackage{hyperref}

\journal{Information Fusion}
\usepackage{etoolbox}
\makeatletter
\patchcmd{\ps@pprintTitle}
  {Preprint submitted to}
  {Accepted for Publication in}
  {}{}
\makeatother

\begin{document}

\begin{frontmatter}


\title{A Comprehensive Overview of Biometric Fusion}


\author{Maneet Singh$^1$, Richa Singh$^1$, Arun Ross$^2$}

\cortext[mycorrespondingauthor]{Richa Singh}
\ead{maneets@iiitd.ac.in, rsingh@iiitd.ac.in, rossarun@cse.msu.edu}
\address{$^1$IIIT Delhi, India}
\address{$^2$Michigan State University, USA}

\begin{abstract}
The performance of a biometric system that relies on a single biometric modality (e.g., fingerprints only) is often stymied by various factors such as poor data quality or limited scalability. Multibiometric systems utilize the principle of \textit{fusion} to combine information from multiple sources in order to improve recognition accuracy whilst addressing some of the limitations of single-biometric systems. The past two decades have witnessed the development of a large number of biometric fusion schemes. This paper presents an overview of biometric fusion with specific focus on three questions: \textit{what} to fuse, \textit{when} to fuse, and \textit{how} to fuse. A comprehensive review of techniques incorporating ancillary information in the biometric recognition pipeline is also presented. In this regard, the following topics are discussed: (i) incorporating data quality in the biometric recognition pipeline; (ii) combining soft biometric attributes with primary biometric identifiers; (iii) utilizing contextual information to improve biometric recognition accuracy; and (iv) performing continuous authentication using ancillary information. In addition, the use of information fusion principles for presentation attack detection and multibiometric cryptosystems is also discussed. Finally, some of the research challenges in biometric fusion are enumerated. The purpose of this article is to provide readers a comprehensive overview of the role of information fusion in biometrics. 
\end{abstract}

\begin{keyword}
\texttt{Biometrics}\sep \texttt{Information Fusion} \sep \texttt{Multibiometrics} \sep \texttt{Soft Biometrics} \sep \texttt{Continuous Authentication} \sep \texttt{Privacy} \sep \texttt{Security} \sep \texttt{Cryptosystems} \sep \texttt{Spoof Detection} \sep \texttt{Social Networks}
\end{keyword}

\end{frontmatter}


\section{Introduction}



Biometrics refers to the automated process of recognizing an individual based on their physical or behavioral traits such as face, fingerprints, voice, iris, gait, or signature~\cite{jain11book}. These traits are often referred to as biometric modalities or biometric cues. Over the past several years, a number of different biometric modalities~\cite{face, iris, fingerprint} have been explored for use in various applications ranging from personal device access systems to border control systems~\cite{jain50years}.

\begin{figure}
\begin{center}
\includegraphics[width=3.4in]{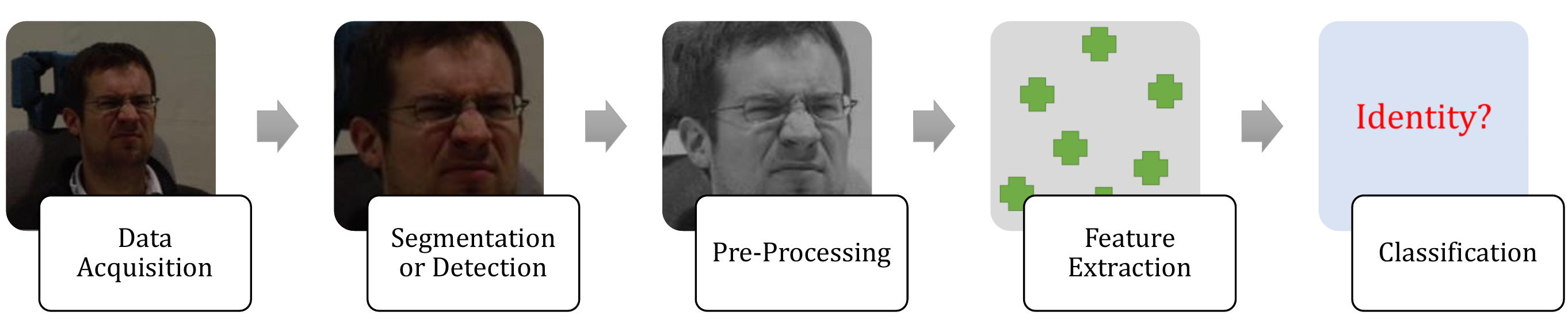}
\caption{General pipeline of a face recognition system. From a pattern recognition perspective, the most significant modules of a biometric system are the sensor module, the segmentation module; the feature extraction module; and the classification or decision-making module.}
\label{fig:pipeline}
\end{center}
\end{figure}

The general framework of a typical biometric recognition system is summarized in Figure \ref{fig:pipeline}. 
Here, given some input data (e,g, an image, video or signal), a typical biometric recognition system first performs segmentation or detection, which involves extracting the modality of interest from the input. This is followed by preprocessing, which involves data alignment, noise removal, or data enhancement. Features are extracted from the preprocessed data, which are then used by a classifier for biometric recognition. The recognition process may involve associating an identity with the input data (e.g., biometric identification) or determining if two instances of input data pertain to the same identity (e.g., biometric verification). 

A {\em unibiometric} system, which utilizes a single biometric cue, may encounter problems due to missing information (e.g., occluded face), poor data quality (e.g. dry fingerprint), overlap between identities (e.g., face images of twins) or limited discriminability (e.g., hand geometry). 
In such situations, it may be necessary to utilize multiple biometric cues in order to improve recognition accuracy. For example, a border control system may use both face and fingerprints to establish the identity of an individual \cite{airportSpain,airportUAE}. In some cases, a biometric cue could be used alongside traditional user-validation schemes such as passwords/passcodes to verify a user's identity. For example, many smartphone devices incorporate such a dual-factor authentication scheme \cite{BiometricsPassCodePatent2012, samsung}. In other applications, multiple sensors could be used to acquire the same biometric modality, thereby allowing the system to operate in different environments. For example, a face recognition system may use both a visible spectrum camera as well as a near-infrared camera to image a person's face and facilitate biometric recognition in a nighttime environment. The aforementioned examples underscore the need for effective biometric {\em fusion} techniques that can consolidate information from multiple sources. 


The term {\em multibiometrics} has often been used to connote biometric fusion in the literature \cite{RossMultibiometricsBook}. In order to develop a multibiometric system, one must consider the following three questions, (i) \textit{what} to fuse, (ii) \textit{when} to fuse, and (iii) \textit{how} to fuse, each of which have been explored in this article.   

\textit{What} to fuse involves selecting the different sources of information to be combined, such as multiple algorithms or multiple modalities. \textit{When} to fuse is answered by analyzing the different levels of fusion, that is, the various stages in the biometric recognition pipeline at which information can be fused. \textit{How} to fuse refers to the fusion method that is used to consolidate the multiple sources of information. 

Given data from a single modality only (say face only), the performance of a recognition system can often be enhanced by incorporating some ancillary information. Incorporating details such as image quality, subject demographics, soft biometric attributes, and contextual meta-data has shown to improve the performance of recognition systems. While recognition performance is a major metric for evaluating biometric systems, it is important to focus on the security (and privacy) aspect of such systems as well. Information fusion is seen as a viable option for securing the biometric templates in a multibiometric system. Cryptosystems based on multiple modalities have been proposed to securely store biometric templates and prevent access to the original data \cite{nagarTifs12}. Biometric systems are also susceptible to spoof attacks. That is, an adversary can impersonate another person's identity by presenting a fake or altered biometric trait and gain unauthorized access. Information fusion can play a major role in the detection and deflection of such malicious activities \cite{marasco12Btas}. Thus, additionally, this paper presents a survey of information fusion techniques along the lines of: (i) biometrics and ancillary information, (ii) spoof (or presentation attack) detection, and (iii) multibiometric cryptosystems. 

\begin{figure}
\begin{center}
\includegraphics[width=3.5in]{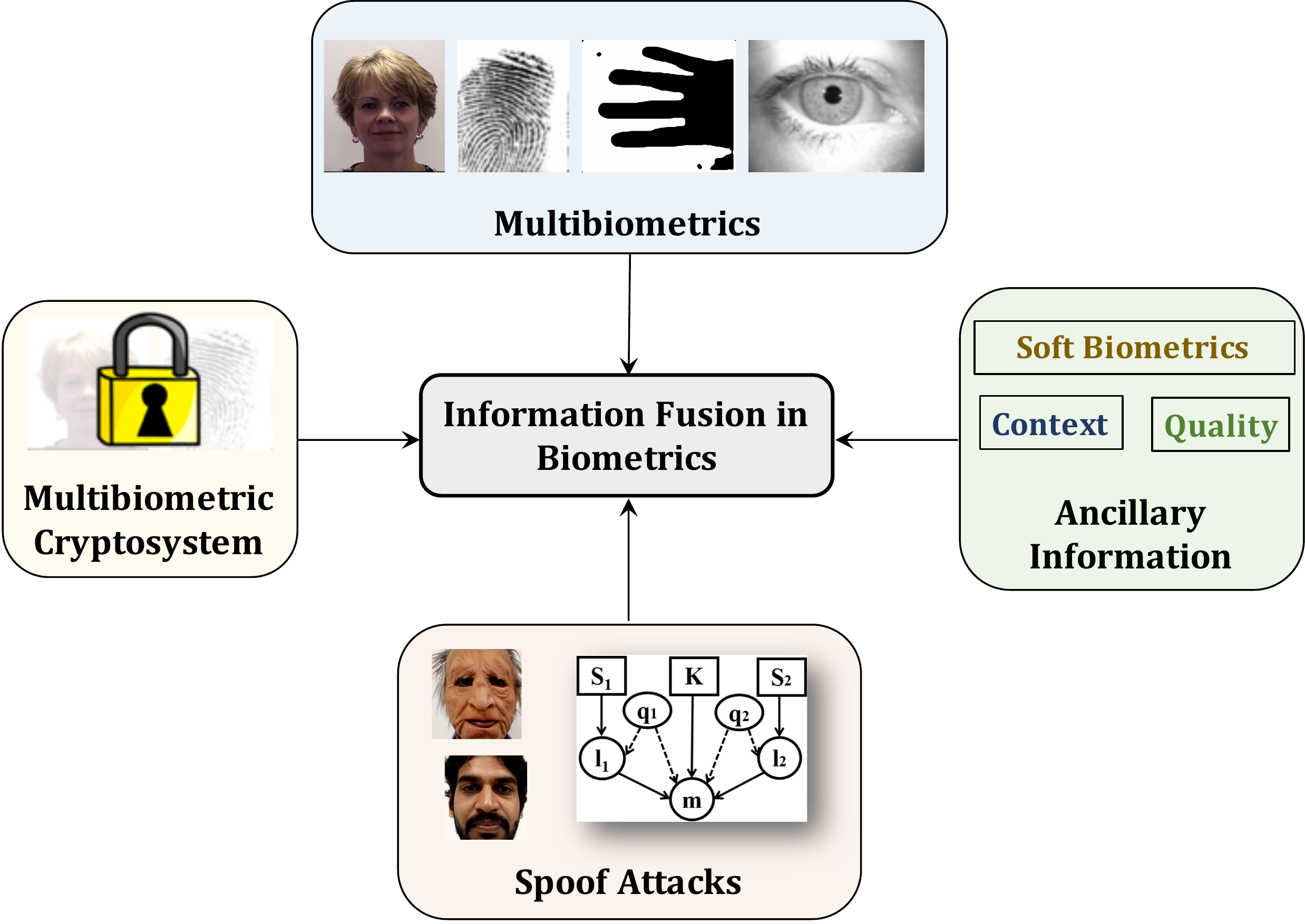}
\caption{Information fusion in the context of biometrics can be used to improve recognition accuracy (multibiometrics, ancillary information) or to improve security (cryptosystems, spoof detection). In some cases, non-biometric cues may also be used in the fusion framework. Images are taken from the Internet, WVU Multimodal dataset \cite{wvuMultimodal}, and MLFP dataset \cite{akshayCvprW}.}
\label{fig:intro}
\end{center}
\end{figure}

\section{Multibiometric Systems}
\label{sec:multi}
A \textit{multibiometric system} can overcome some of the limitations of a unibiometric system by combining information from different sources in a principled manner. The utilization of multiple sources often results in improved recognition performance and enhanced system reliability, since the combined information is likely to be more distinctive to an individual compared to the information obtained from a single source.

\begin{figure*}
\begin{center}
\includegraphics[width=6in]{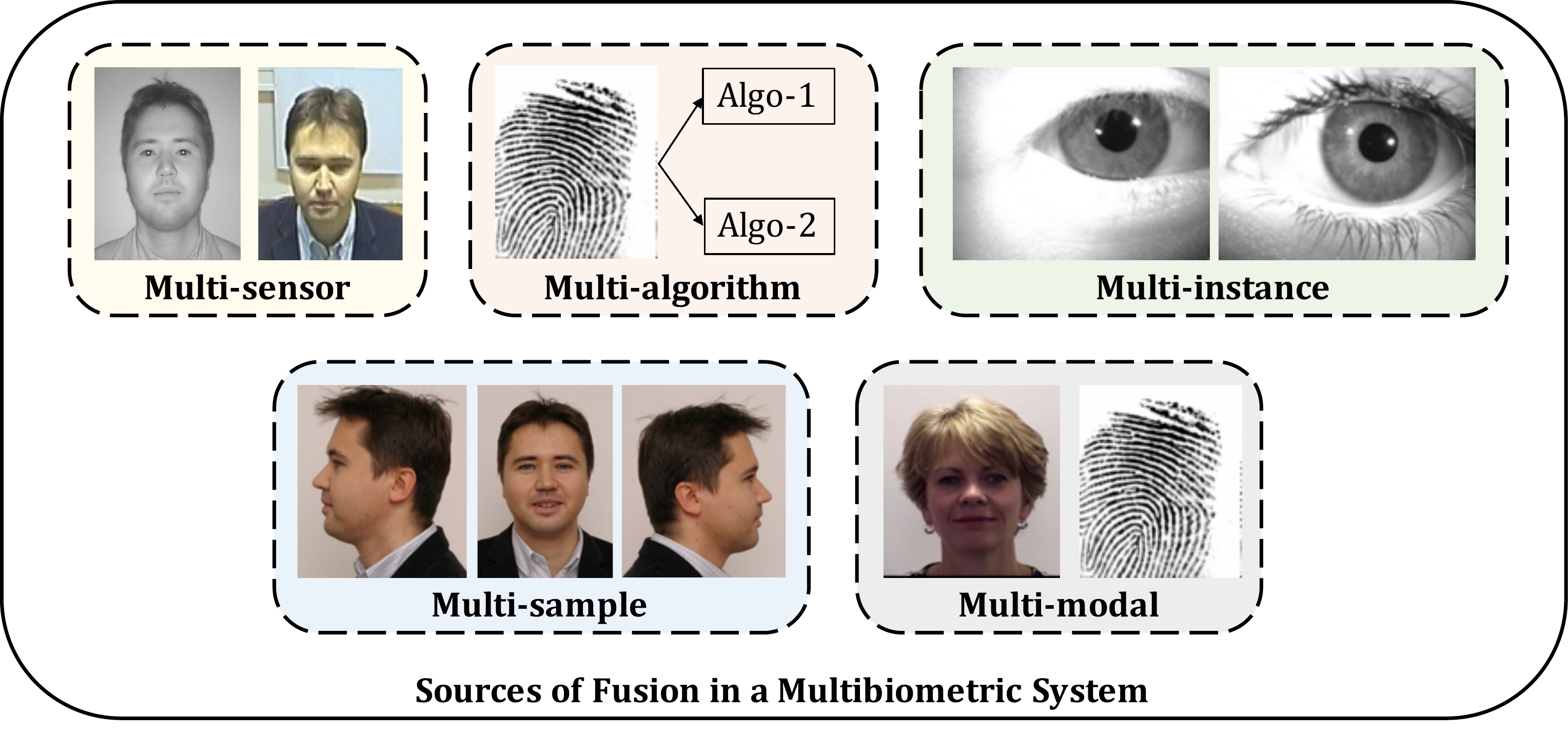}
\caption{Different sources of information that can be exploited by a multibiometric system. Information from multiple sensors (infrared and visible spectra) or multiple algorithms (minutiae-based and texture-based) or multiple instances (left and right irides) or multiple samples (left, frontal and right facial profiles) or multiple modalities (face and fingerprint) can be fused.}
\label{fig:source}
\end{center}
\end{figure*}

\subsection{Sources of Fusion}
As mentioned previously, one of the major questions for developing a multibiometric system is \textit{what} to fuse. Figure \ref{fig:source} presents the different sources of information that can be fused in a multibiometric system. Depending upon the sources of fusion, a multibiometric system can correspond to one of the following configurations: (i) multi-sensor, (ii) multi-algorithm, (iii) multi-instance, (iv) multi-sample, or (v) multi-modal.
\vspace{3pt}

\noindent \textbf{(i) Multi-sensor} systems combine information captured by multiple sensors for the same biometric modality. For example, a face recognition module could utilize RGB data captured using a visible spectrum camera, along with depth information captured using a 3D camera \cite{goswami14Tifs} or infrared data captured using an NIR camera \cite{cvejic07Cvpr,singh08Pr,singh08If}. Using both the images for identifying a subject would result in a multi-sensor fusion algorithm. Such systems rely on a single modality for recognition; however, they capture different information from the same modality by utilizing multiple sensors \cite{BourlaiBook}. They are useful in scenarios which require a different mode of capture at different times, or where discriminative information can successfully be captured by different sensors. 
\vspace{3pt}

\noindent \textbf{(ii) Multi-algorithm} systems utilize multiple algorithms for processing an input sample. Data is captured from a biometric modality using a single sensor; however, multiple algorithms are used to process it. For example, a fingerprint recognition system could utilize both minutiae and texture features for matching fingerprints \cite{RossHybrid03}, or a palmprint recognition system could utilize Gabor, line, and appearance based palmprint representations for matching \cite{kumar05PR}. Such systems benefit from the advantage of extracting and utilizing different types of information from the same sample. In cases where two algorithms or feature sets provide complementary information, multi-algorithm systems can often result in improved performance.\vspace{3pt}

\noindent \textbf{(iii) Multi-instance} systems capture multiple instances of the same biometric trait. In the case of iris recognition, the recognition module can utilize both left and right irides, thereby resulting in a multi-instance system \cite{rathgebCs14}. Similarly, in the case of a fingerprint or palm-print recognition system, a multi-instance system can utilize data captured from the ten fingers or both palms \cite{nandakumarBtas08,uhl09}. Multi-instance systems may use the same feature extraction and matching methods for all instances of the biometric trait.
\vspace{3pt}

\noindent \textbf{(iv) Multi-sample} systems work with multiple samples of the same biometric modality, often captured with some variations. Video-based recognition models fall under this category, where, a biometric modality is captured continuously over a small period of time (e.g., several seconds long). This often results in a large number of frames containing multiplicity of information \cite{ParkFaceVideo2007}. In the literature, videos have been used for performing face \cite{ParkFaceVideo2007,goswami17Tifs,bhatt14Tifs,zhou03Cviu} and gait \cite{wang04Tran,hu13Tran,nixon10gait,kale04Tip} recognition. This results in a multi-sample recognition system, which combines information captured across multiple video frames. Such systems are able to extract diverse information from a single biometric modality, while requiring only a single sensor.
\vspace{3pt}

\noindent \textbf{(v) Multi-modal} systems utilize data captured from multiple biometric cues in order to recognize a subject. A multi-modal system could utilize information captured from face, fingerprint, and iris modalities \cite{goswami16If}; face, fingerprint, and speech modalities \cite{jain99multimodal}; face and voice modalities \cite{ben99TNN,msuavis2018}; ear and face modalities \cite{chang03Pami}; or even iris and periocular modalities \cite{woodard10Icpr,zhang18Tifs}. Research has also focused on combining different modalities for performing speaker recognition, such as audio and lip motion \cite{wark01Dsp}; audio, lip motion and lip texture \cite{cetingul06Sp}; and audio, RGB and depth information \cite{minotto15TMM}. Such systems can eliminate the limitations of a particular biometric modality by having the flexibility of processing multiple modalities \cite{fierrez18IF,kumar16Fusion}. Multi-modal systems are also useful in scenarios where an individual cannot provide data for a particular biometric modality (say injured fingerprints), but can provide data pertaining to another one (say face). Fusing information from different modalities further enables extraction of distinctive features, often resulting in enhanced recognition performance \cite{rossPrl}. 

Besides the aforementioned sources, non-biometric cues may also be used in the fusion process. As will be described later, information such as contextual meta-data can be used in conjunction with biometric identifiers in order to recognize an individual \cite{bharadwajIjcb14}.

\subsection{Levels of Fusion}
\begin{figure}
\begin{center}
\includegraphics[width=3.4in]{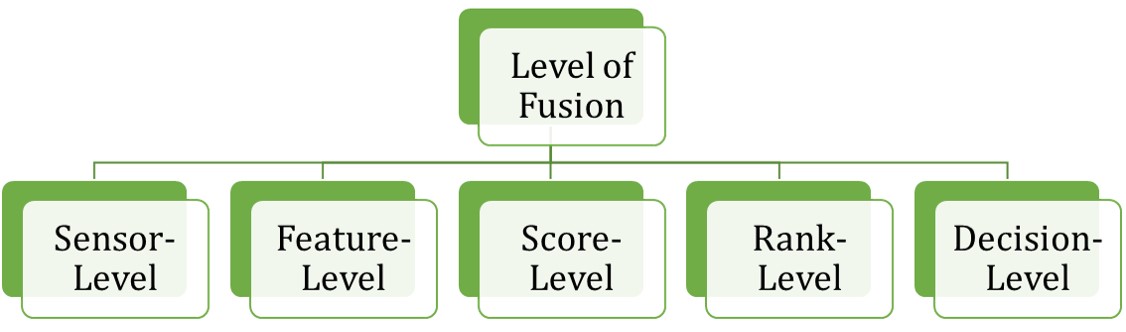}
\caption{Levels of fusion in a multibiometric system. These levels correspond to the various modules of a typical biometric system. See Figure \ref{fig:pipeline}. While these levels of fusion are applicable to both verification and identification systems, rank-level fusion is typically applicable to only identification systems.}
\label{fig:level}
\end{center}
\end{figure}

Figure \ref{fig:level} presents the different levels at which fusion can be incorporated in a biometric pipeline, viz., (i) sensor-level, (ii) feature-level, (iii) score-level, (iv) rank-level, or (v) decision-level. Each of these levels of fusion are explained in detail below.
\vspace{3pt}

\noindent \textbf{(i) Sensor-level} fusion or data-level fusion generally corresponds to multi-sensor or multi-sample algorithms, where data is combined immediately after its acquisition. That is, data fusion is carried out prior to feature extraction, directly on the \textit{raw} data~\cite{OthmanRossMixing2012}. In case of a face recognition module, this corresponds to direct pixel-level combination of face images captured from a camera. For example, multiple faces can be captured with pose variations such as frontal, left profile, or right profile. A mosaicing technique can be used to fuse the samples together in order to obtain a combined face representation \cite{singh07Tran}. Often a direct fusion strategy or adding pixels of two images is utilized \cite{RossImageFeatureMosaic06}. 
\vspace{3pt}

\noindent \textbf{(ii) Feature-level} fusion refers to algorithms where fusion is performed on multiple features extracted from the same or different input data. This could correspond to multiple feature sets pertaining to the same biometric trait, such as textural and structural features of a face image or different features from a hand or palm-print image \cite{kong06Pr,kumar06Tip}. It could also correspond to features extracted from different modalities, such as face and hand images \cite{ross05Spie}. Such algorithms are often used by multibiometric cryptosystems, where features from multiple biometric sources are combined to improve security and privacy \cite{nagarTifs12}. They have also been used for indexing multimodal biometric databases~\cite{GyaourovaIndexCodes2012}. Feature-level fusion combines different representations in order to generate a single representation for a given individual. For example, representation learning algorithms can be used to learn a shared representation of features extracted from different modalities \cite{SharedRepStanLi2015}.
\vspace{3pt}

\noindent \textbf{(iii) Score-level} fusion corresponds to algorithms where the match scores produced by different matchers are fused together. Some of the common fusion algorithms applied at this level are mean score fusion, max score fusion, or min score fusion, where the mean, maximum, or minimum score of multiple matchers is considered as the final score \cite{JainScoreNorm2005, rossPrl,pohPR06,hePR10,yilmaz16Fusion}. Dempster-Shafer theory or probabilistic techniques such as likelihood ratio based score fusion have also been applied in the literature \cite{pohBtas07,nandakumarPami08,vatsa07Jns,poh12Pami}. In addition, Ding and Ross \cite{DingRossImputation2012} discuss several imputation techniques for handling missing or incomplete information in the context of score-level fusion. This is the most common type of fusion described in the literature due to the ease of accessing scores generated by commercial matchers. Most commercial matchers do not provide easy access to features or, sometimes, even the raw data. 
\vspace{3pt}

\noindent \textbf{(iv) Rank-level} fusion is performed after comparing the input probe with the templates in the gallery set, i.e., the database. In the task of identification, where, a given probe image is compared against a gallery of images, a ranked list of matching identities is often generated by the matcher. In the literature, the rank lists from multiple matchers have been fused using techniques like Borda count, logistic regression, and highest rank method \cite{abazaBtas09,tinPami94,monwar09Tran,kumar11Sys,bharadwajIjcb14}. In scenarios having limited access to features or match scores, rank-level fusion is often deemed effective.
\vspace{3pt}

\noindent \textbf{(v) Decision-level} fusion corresponds to algorithms where fusion is performed at the decision level \cite{monwar09Tran,veeramachaneni08Cvprw,prabhakar02Pr}. Majority voting is one of the most common fusion algorithms applied at the decision level. Decisions taken by $n$ matchers or classifiers are combined based on a majority vote, resulting in a final decision. Decision-level fusion has the advantage of working well with black-box systems, where only the final decisions are available \cite{indovina03}. This is true in the case of many commercial systems, where access to features, scores and ranks may not be feasible.

Thus, fusion in biometrics can be invoked at different levels in the biometric recognition pipeline and can avail of different sources of information. For a detailed review on the sources and levels of fusion, the reader is encouraged to refer to \cite{RossMultibiometricsBook,ross09multibiometric}. 

The final piece for developing a multibiometric framework is understanding \textit{how} to fuse the diverse sources of information. The remainder of this paper presents an expansive survey of information fusion techniques applied in the context of (i) combining biometrics and ancillary information, (ii) spoof detection, and (iii) designing multibiometric cryptosystems.

\begin{figure}
\begin{center}
\includegraphics[width=3in]{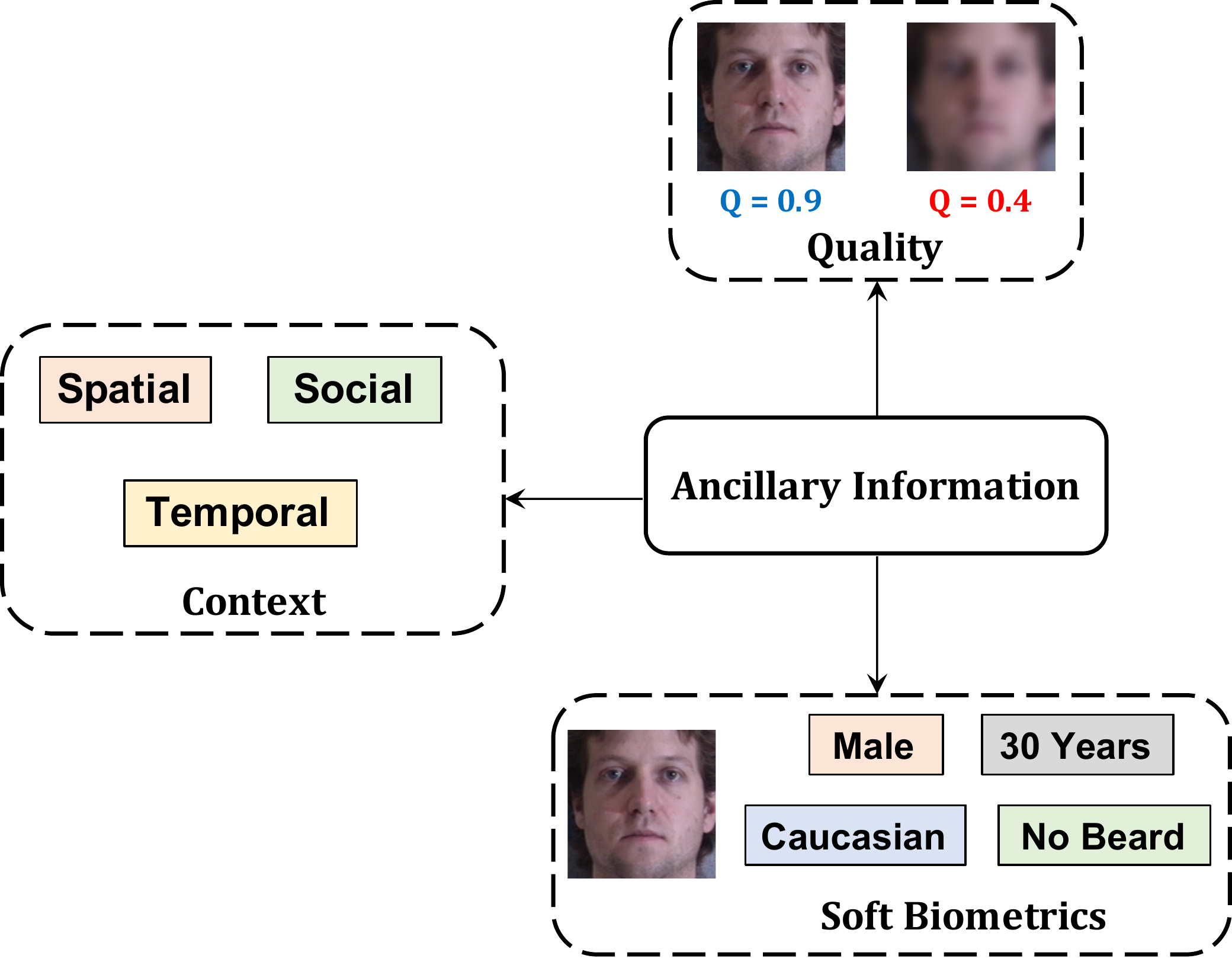}
\caption{Types of ancillary information that can be combined with primary biometric traits (such as faces and fingerprints) in order to improve recognition accuracy. Data quality, soft biometric attributes and contextual information can aid in the biometric recognition process.}
\label{fig:ancillary}
\end{center}
\end{figure}

\section{Biometrics and Ancillary Information}
Researchers have incorporated ancillary information in the traditional biometrics pipeline in order to improve recognition performance. Ancillary data refers to any additional information that can be provided about a particular biometric sample which might aid in the recognition process. Figure \ref{fig:ancillary} presents some of the commonly used sources of ancillary information, viz., quality estimates, soft biometric attributes, and contextual information. Ancillary information has also been used to perform continuous authentication of a subject. This section presents an overview of the literature associated with each form of ancillary information mentioned above. 

\begin{figure*}
\begin{center}
\includegraphics[width=5in]{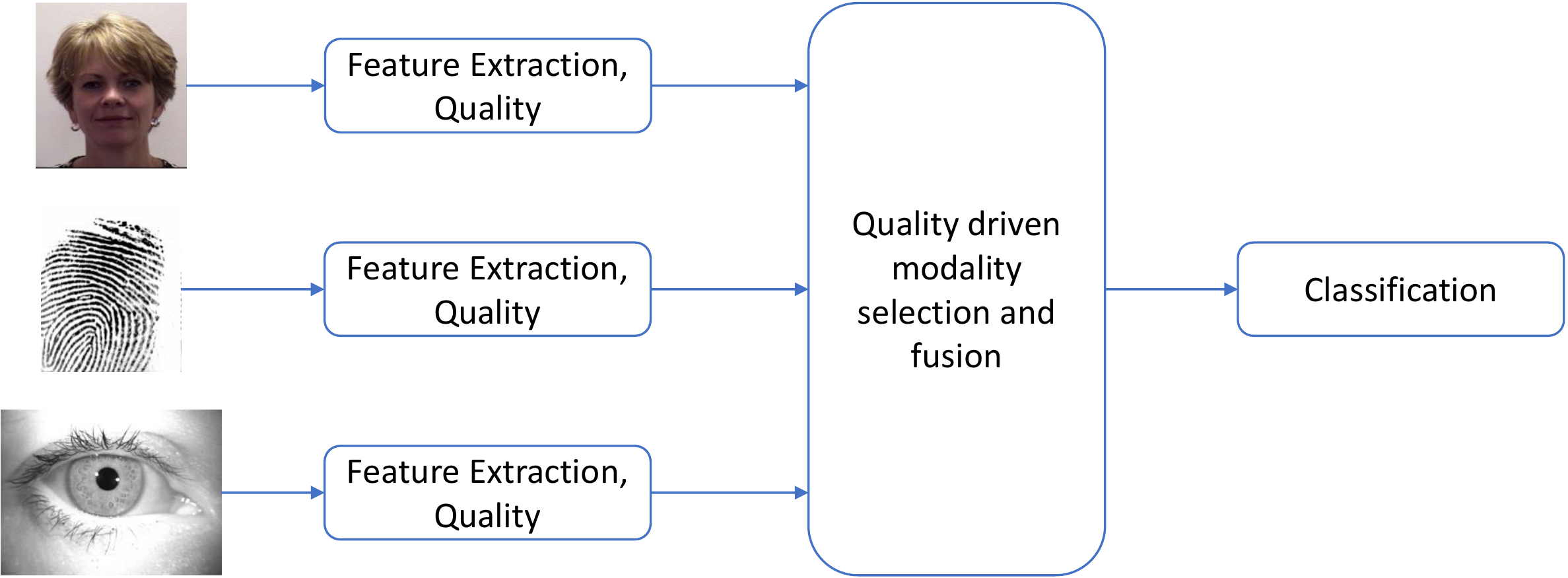}
\caption{Given input from multiple modalities, quality information is often used for modality selection, fusion, or context switching at run-time.}
\label{fig:contextSwitching}
\end{center}
\end{figure*}

\begin{figure}
\begin{center}
\includegraphics[width=3.2in]{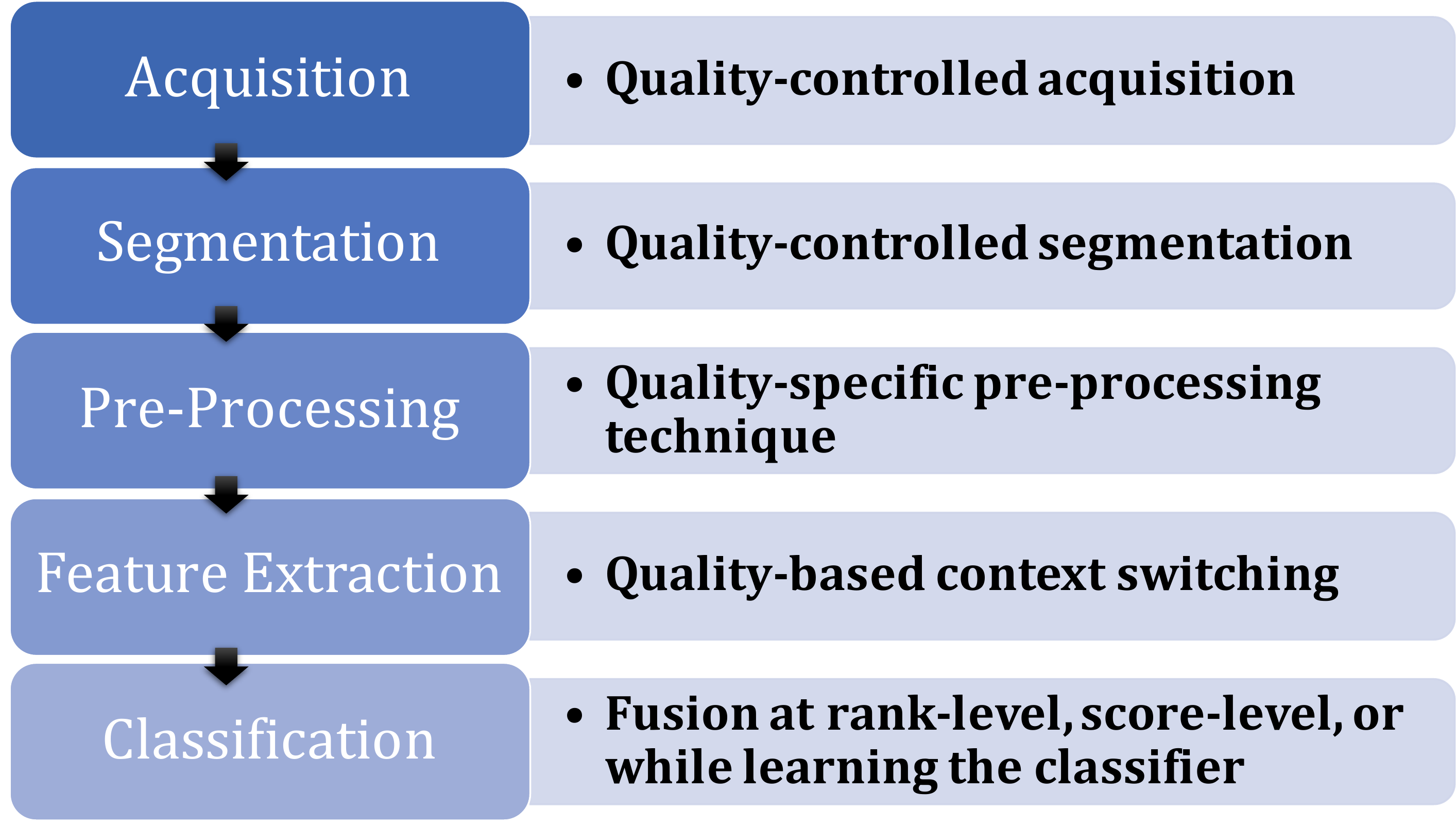}
\caption{Biometric sample quality has been incorporated into various modules of a biometric system. In addition, a number of fusion rules have been proposed to combine quality with features, match scores and ranks.}
\label{fig:quality}
\end{center}
\end{figure}

\subsection{Biometrics and Quality}
As per ISO standards (ISO/IEC 29794-1), a biometric sample is said to be of good quality if ``it is suitable for automated matching''. For a biometric sample, the quality is often quantified by the ease with which an image can be processed, including feature extraction and correct classification with a high confidence score. A good quality biometric sample is often associated with rich features and easy classification, whereas a poor quality sample suffers from the inherent challenge of noisy data. Bharadwaj \textit{et al.} \cite{bharadwaj14Eurasip} present a review of biometric quality for the face, fingerprint, and iris modalities. A comprehensive survey of different quality measures proposed in the literature is presented, along with their estimation strategies and methods of incorporating quality in the biometric classification pipeline. Experimental analysis on a multimodal biometric dataset reiterated the importance of carefully selecting quality measures for enhancing recognition performance. In the literature, a sample's quality estimate has been used as ancillary information in both unibiometric and multibiometric recognition systems. As shown in Figure \ref{fig:quality}, this is achieved by exploiting the quality information at different stages in the recognition pipeline. Figure \ref{fig:contextSwitching} demonstrates the inclusion of quality information for modality selection, fusion, or context switching in multi-modal recognition scenarios.   

In 2003, Bigun \textit{et al.} \cite{bigun03} proposed incorporating quality information into a Bayesian statistical model for performing multimodal biometric classification. Quality is incorporated as a variance parameter  such that samples with higher quality are associated with lower variance. The entire framework consists of two \textit{supervisors}: client and impostor, that are trained for performing verification. Fierrez-Aguilar \textit{et al.} \cite{julianPR04} built upon the above architecture and presented one of the earliest works in the literature on fusing biometric quality at the score level, for performing multimodal biometric authentication. The proposed algorithm is built over an SVM, where the training function is modified to include a quality-based cost term, thereby associating more weight with higher quality training samples. At the time of testing, the scores generated for each modality are combined in a weighted manner, where the weights are dependent on the quality scores. This ensures that samples with higher quality are given more weight when performing score level fusion of multiple modalities. Poh and Bengio \cite{poh05} introduced a confidence criterion to incorporate quality when combining multiple biometric classifiers in a linear manner. Quality is considered as the derived margin, i.e. the difference between the False Acceptance Rate and the False Rejection Rate. This quality measure is integrated as an \textit{a priori} weight when performing fusion of multiple classifiers.   
 
In an attempt to understand the impact of fingerprint quality on different classifiers, Fierrez-Aguilar \textit{et al.} \cite{aguilar05} performed experiments using ridge-based and minutiae-based classifiers. Fingerprint images having different quality scores were used for testing and it was observed that the ridge-based system was more robust to quality variations. A weighted adaptive score fusion technique was also proposed, which combines the scores obtained from the ridge-based system ($s_R$) and minutia-based system ($s_M$) as follows:
\begin{equation}
s_Q = \frac{Q}{2} s_M + \bigg(1 - \frac{Q}{2}\bigg)s_R.
\end{equation}
Here, $Q$ refers to the quality of the image, and $s_Q$ refers to the final score generated by the entire framework. Consistent with their findings, as the image quality decreases, more weight is given to the score generated by the ridge-based classifier. 

In 2006, Nandakumar \textit{et al.} \cite{nandakumar06} proposed the use of a single quality metric for both the template (image stored in the gallery database) and the probe (image presented during verification). The authors proposed a Quality-based Product Fusion Score (QPFS) which is based upon the joint density of the match score and quality estimated from a given gallery-probe pair:
\begin{equation}
QPFS(s) = \prod_{j=1}^{R} \frac{l_{j, gen}(s_j, q_j)}{l_{j,imp} (s_j, q_j)}, 
\end{equation}
where, $l_{j,gen} (s_j, q_j)$ refers to the joint density of the match score ($s_j$) and quality ($q_j$) of the $j^{th}$ sample. In 2008, they further built upon the proposed likelihood ratio based technique by estimating the joint density using Gaussian Mixture Models \cite{nandakumarPami08}. They tested their approach on a bimodal system involving fingerprint and iris. 

\begin{table*}
\centering
\caption{A brief summary of techniques incorporating biometric sample quality in the biometric recognition pipeline.}
\label{tab:quality}
\begin{tabular}{|c|l|l|}
\hline
\textbf{Year} & \textbf{Authors} & \textbf{Description} \\
\hline
\hline
2003 & Bigun \textit{et al.} \cite{bigun03} & Quality is incorporated in a Bayesian model as a variance parameter \\
\hline
2005 & Fierrez-Aguilar \textit{et al.} \cite{julianPR04} & SVM-based model which incorporates quality in its cost function \\
\hline
2005 & Poh and Bengio \cite{poh05} & Margin based quality measure is incorporated as an \textit{a priori} weight for score fusion \\
\hline
2005 & Fierrez-Aguilar \textit{et al.} \cite{aguilar05} & Weighted score fusion technique for fingerprint recognition using different features \\
\hline
2006 & Nandakumar \textit{et al.} \cite{nandakumar06} & Quality-based Product Fusion Score (QPFS) utilizing quality score for gallery and probe pair \\
\hline
2007 & Poh \textit{et al.} \cite{pohBtas07} & Score normalization technique incorporating device and quality information \\
\hline
2007 & Vatsa \textit{et al.} \cite{vatsa07Jns} & RDWT-based quality utilized for multimodal recognition using $2\nu$-SVM fusion algorithm \\
\hline
2008 & Nandakumar \textit{et al.} \cite{nandakumarPami08} & Built over QPFS model by incorporating Gaussian Mixture Models \\
\hline
2008 & Maurer and Baker \cite{maurerPr08} & Bayesian Belief Network modeling \textit{local} and \textit{global} quality \\
\hline
2009 & Poh \textit{et al.} \cite{pohTifs09} & Benchmarked existing multimodal fusion algorithms under varying quality and cost \\
\hline
2009 & Abaza and Ross \cite{abazaBtas09} & Q-based Borda Count technique incorporating quality estimate at rank-level fusion \\
\hline
2009 & Vatsa \textit{et al.} \cite{vatsa09Jar} & Quality-augmented fusion technique of level-2 and level-3 fingerprint features\\
\hline
2010 & Poh \textit{et al.} \cite{pohTifs10} & Incorporates score normalization \cite{pohBtas07} as pre-processing for different multimodal pipelines \\
\hline
2010 & Tong \textit{et al.} \cite{tongCvprw10} & Bayesian Belief Network which incorporates quality estimates of probe and gallery image \\
\hline
2010 & Vatsa \textit{et al.} \cite{vatsa10Icpr} & Quality driven image and score level fusion of iris images, followed by probabilistic SVM  \\
\hline
2010 & Vatsa \textit{et al.} \cite{vatsa10Tifs} & Quality driven dynamic context switching for classifier or fusion selection \\
\hline
2012 & Zhou \textit{et al.} \cite{zhouSmc12} & Eye recognition is performed using quality estimates of segmented regions \\
\hline
2013 & Rattani \textit{et al.} \cite{RattaniPohRoss2013} & Incorporating sensor influence on image quality and match scores using a graphical model \\
\hline
2015 & Bhardwaj \textit{et al.} \cite{bharadwaj15Pr} & \textit{QFuse}: An online learning framework utilizing quality based context switching \\
\hline
2015 & Huang \textit{et al.} \cite{huangPr15} & Adaptive Biomodal Sparse Representation based Classification - quality with Sparse Coding \\
\hline
2016 & Ding \textit{et al.} \cite{DingBBN_ICB2016} & Multiple Bayesian Belief Models for fusing match scores and quality values \\
\hline
2016 & Muramatsu \textit{et al.} \cite{muramatsu16} & View  Transformation Model with quality-based score normalization\\
\hline
2017 & Liu \textit{et al.} \cite{quality17Cvpr} & Quality Aware Network (QAN): Fuse predicted quality score in a CNN \\
\hline
\end{tabular}
\end{table*}

Another likelihood-ratio based algorithm was proposed by Poh \textit{et al.} \cite{pohBtas07}, where the authors proposed incorporating both the device information and the quality information for predicting the class label of an input sample. Such a technique can be utilized in scenarios where data is collected from multiple sensors for a particular modality. Since in real world scenarios, the acquisition device can be unknown at the time of testing, a posterior probability is estimated using the quality measures. The proposed score normalization technique is then written as,
\begin{equation}
y^{norm} = \log \frac{\sum_d p(y|C, d) p(d|q)}{\sum_d[(y|I, d) p(d|q)]},
\end{equation}
where, $C$ and $I$ refer to the client and impostor classes, respectively, and $d$, $q$ corresponds to the device and quality estimate of the given sample, respectively. $y$ refers to a vector of scores generated by different classification devices. A major limitation of this algorithm is the difficulty in keeping track of the number of devices that can be used for generating the probe images, thus limiting the applicability of the proposed model. In 2010, this model was further extended \cite{pohTifs10} and experiments were performed by incorporating quality-based score normalization as a pre-processing step in existing multi-modal fusion pipelines. Experimental analysis under different situations (known or unknown device) demonstrated the efficacy of the proposed model. Vatsa \textit{et al.} \cite{vatsa07Jns} proposed computing the quality score of a given biometric image using Redundant Discrete Wavelet Transform (RDWT). Results were shown for the task of multimodal recognition of face and iris images. The independent match scores obtained were multiplied by the quality scores and fused using a novel 2$\nu$-SVM fusion algorithm.

Maurer and Baker \cite{maurerPr08} presented a detailed description and analysis of a Bayesian Belief Network that was proposed earlier by them \cite{baker05fusion}. The model is built upon the motivation that ``\textit{if the match score (similarity score) of a low quality sample is high, it is extremely unlikely to be from an impostor}''. This implies that simply providing a weight based on the quality of the sample while performing multi-modal fusion might negatively impact some true positive (genuine) samples. In order to address this, the authors proposed a Bayesian Belief Network where no dependence is encoded between the quality and identity. The model is used for performing multimodal identification, or even identification of a single modality with multiple samples. The proposed architecture encodes quality in the form of \textit{local} and \textit{global} measures, where the global quality brings together the individual (local) multiple qualities. 

Owing to the importance of multimodal biometric authentication in real world scenarios and the need for establishing a better understanding of different approaches, in 2009, Poh \textit{et al.} \cite{pohTifs09} benchmarked several multimodal fusion algorithms. The algorithms were evaluated in terms of their recognition performance under different quality and cost constraints. The evaluation was carried out on the first-of-its-kind BioSecure DS2 dataset, consisting of data pertaining to several modalities, along with their quality estimates. One of the key observations of their experiment was that the fusion algorithms which incorporate derived quality measures at run-time for cross-device experiments provided better results, compared to techniques that ignore quality measures. Abaza and Ross \cite{abazaBtas09} presented a Q-based Borda Count technique, which incorporates the quality estimate of the input probe image and gallery sample at the rank-level. The authors proposed modifying the Borda Count technique by incorporating the quality as a weight, in order to reduce the contribution of bad quality samples while performing rank-level fusion. The Q-based Borda Count technique can be written as:
\begin{equation}
R_j = \sum_{i=1}^{C} Q_{i,j}r_{i,j}
\end{equation}
where, $R_j$ refers to the final fused rank corresponding to the $j^{th}$ subject of the gallery with respect to $C$ biometric classifiers. $Q_{i,j}$ is the minimum quality value between the $j^{th}$ user's gallery image and the $i^{th}$ probe. $r_{i,j}$ refers to the rank assigned to the $j^{th}$ user of the database by the $i^{th}$ classifier. It was observed that including the sample's quality in a multiplicative manner ensures less weight to low quality samples. Since the fusion techniques are applied at the rank-level, they can easily be incorporated into any existing multimodal classification system. Vatsa \textit{et al.} \cite{vatsa09Jar} proposed a quality-augmented fusion technique of level-2 and level-3 features of fingerprints, based on the Dezert-Smarandache (DSm) theory of paradoxical reasoning. The authors address the scenario of missing information of low quality fingerprint images captured in real world scenarios. Quality scores are computed using the RDWT technique described earlier \cite{vatsa07Jns}, followed by the extraction of level-2 and level-3 features of the given fingerprint image. These features are then augmented by the quality measure and fused using the DSm theory to obtain a final match score. 

Tong \textit{et al.} \cite{tongCvprw10} built upon the model proposed by Maurer and Baker \cite{maurerPr08} and proposed a Bayesian Belief Network which incorporates the quality estimates of the probe and the gallery images. They emphasized that while it is assumed that all gallery images would have higher quality, it must nevertheless be incorporated when performing identification, along with the quality of the probe sample. The causal relationships between the quality of samples, the decision (match/non-match) and the scores are modeled via a probabilistic graphical structure.

Working with a single modality, Vatsa \textit{et al.} \cite{vatsa10Icpr} improved the performance of iris recognition by calculating RGB channel-based quality scores and performing fusion using the two lowest quality channels. Match scores corresponding to the fused image and the highest quality image are then provided as input to a probabilistic support vector machine for obtaining the final fused match score. The proposed framework based on image-level and score-level fusion was shown to achieve improved performance. Further, they also presented a classification framework which utilized the quality vector for performing context switching to dynamically select the appropriate fusion or classification algorithm at run-time\cite{vatsa10Tifs}. In 2012, Zhou \textit{et al.} \cite{zhouSmc12} proposed a quality driven technique for performing eye recognition. The algorithm involved segmenting the eye into iris and sclera, followed by computing their quality estimates independently. Independent models are trained on all three regions of the eye and, based on the quality of the three regions, a single region is selected;  classification is then performed using only this selected region. Ding \textit{et al.} used Bayesian graphical models to understand the impact of various variables on image quality and vice-versa, and developed techniques to fuse quality with match scores and liveness values in a fingerprint verification system \cite{DingBBN_ICB2016}.

Bharadwaj \textit{et al.} \cite{bharadwaj15Pr} proposed \textit{QFuse}, an online learning framework incorporating quality based context switching for multimodal recognition. Multiple quality metrics are estimated for a given gallery and probe pair, which are provided as input to an ensemble of SVMs for choosing between unimodal classification or fusion-based classification. The incorporation of online learning in the QFuse framework makes it more applicable and usable in real world scenarios. Huang \textit{et al.} \cite{huangPr15} proposed Adaptive Bimodal Sparse-Representation based Classification (ABSRC) for performing feature fusion based on the quality of two samples. A two-step framework is proposed, where initially independent dictionaries are learned for both the modalities. Based on these dictionaries, the feature vector and Sparse Coding Error (SCE) are calculated for the samples, which are then used as quality measures. The SCE is used to generate weights, based on which the two feature vectors of different modalities are concatenated. A similar approach is followed for the learned dictionaries as well, where the dictionaries are simply concatenated based on the weights obtained. The second stage dictionary is used for performing classification. Muramatsu \textit{et al.} \cite{muramatsu16} proposed a view transformation model which incorporates quality measure for cross-view gait recognition. Quality values are used to develop a score normalization framework for recognizing gait samples from different views. 

\begin{figure*}
\begin{center}
\includegraphics[width=5in]{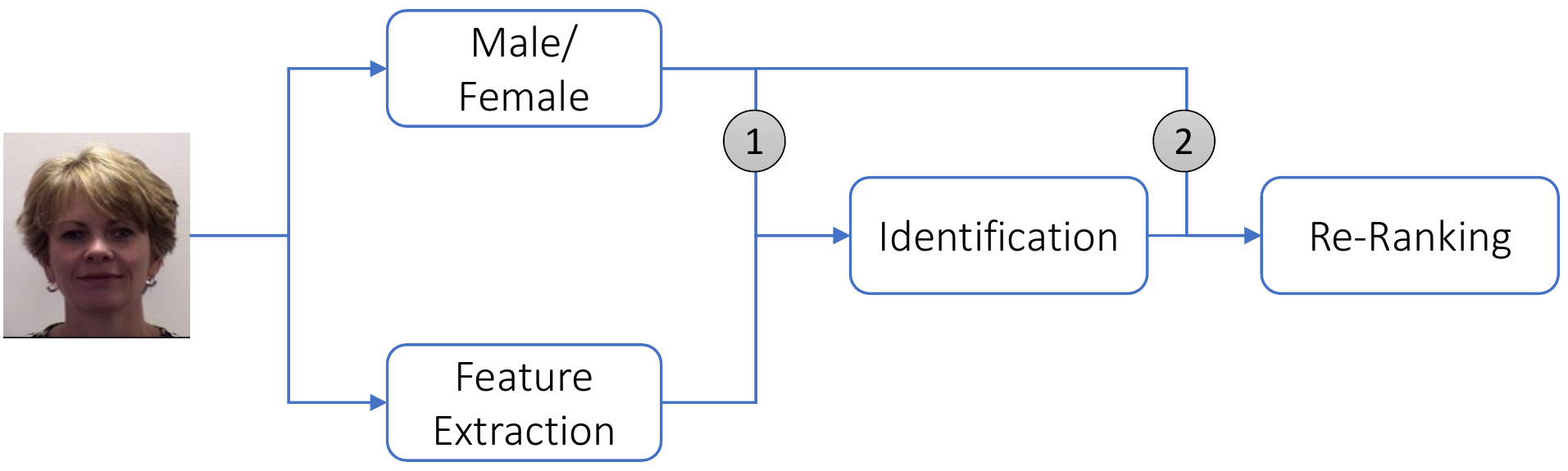}
\caption{Example of soft biometric fusion in a general biometric recognition pipeline. (i) Soft biometric information can be fused with primary biometric features for classification, or (ii) soft biometric information can be used for re-ranking the identification list obtained from primary biometric traits.}
\label{fig:softFusion}
\end{center}
\end{figure*}

It can thus be observed that quality based fusion has been performed at the feature level, score level, and decision level, by different algorithms. Research began with several probabilistic models being proposed for incorporating quality in the recognition framework; however, recent advances have led to the development of representation learning based models incorporating quality estimates in the learning stage as well. Quality information has also been utilized to create context switching based algorithms which select an algorithm for processing the input sample based on its quality. It is interesting to observe from Table \ref{tab:quality} that research at the intersection of quality and biometrics has seen some decline in recent literature. Recently, Liu \textit{et al.} \cite{quality17Cvpr} proposed Quality Aware Networks (QANs), for performing set-to-set matching, with application in person re-identification. QAN is built over Convolutional Neural Networks (CNNs), and strengthens our hypothesis that representation learning techniques including deep learning could benefit from fusing quality information in the recognition pipeline.  Improved performance might be attained for challenging problems when matching images across scenarios such as cross-resolution face recognition or cross-sensor fingerprint matching.

\begin{figure}
\begin{center}
\includegraphics[width=3.2in]{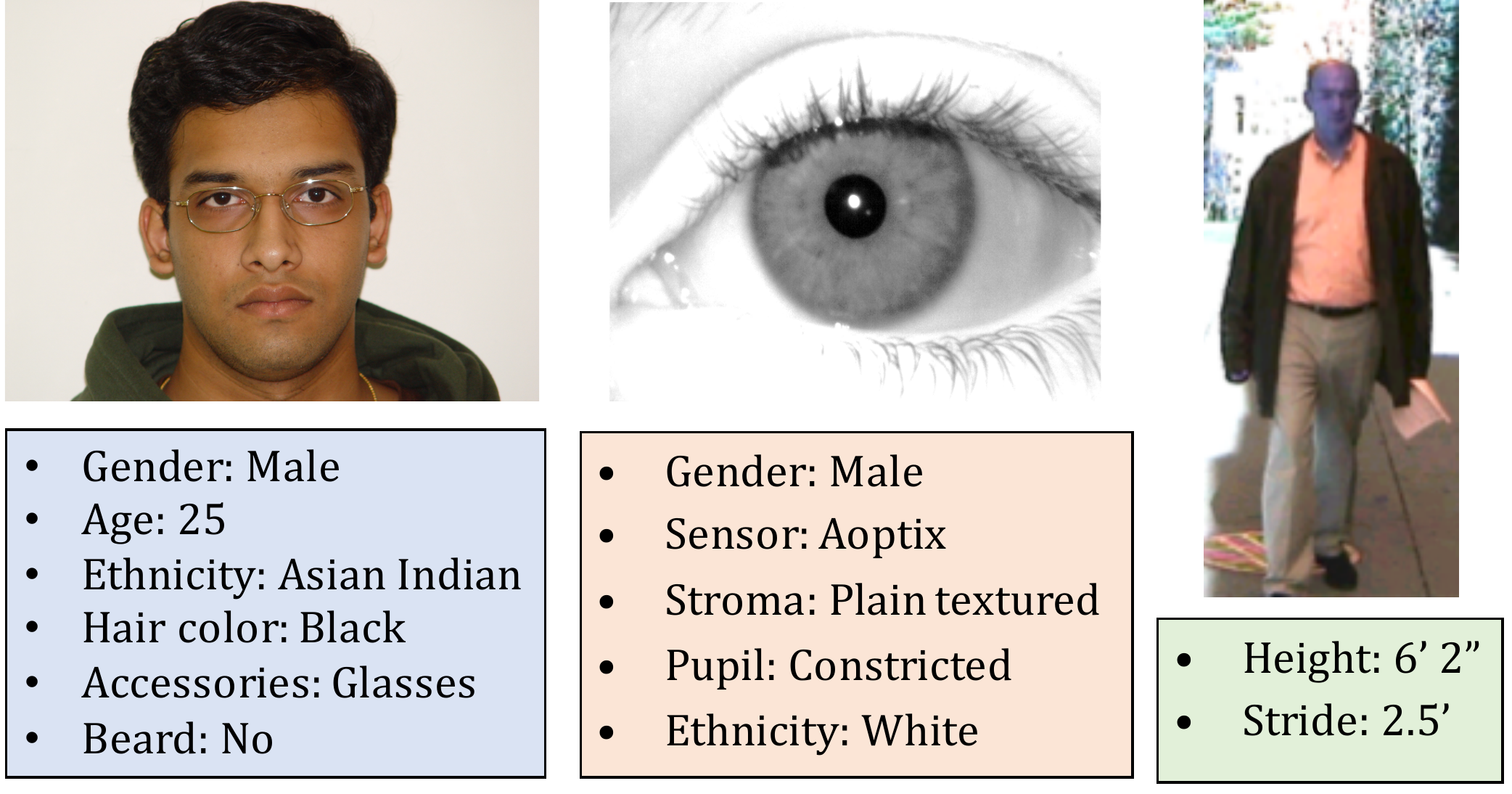}
\caption{Examples of soft biometric information evident in the face, iris and gait modalities. In some cases, manually labeled or annotated soft biometric attributes are used. In other cases, these attributes are automatically extracted.}
\label{fig:soft}
\end{center}
\end{figure}

\subsection{Primary Biometrics and Soft Biometrics}    
Soft biometrics refer to the ``\textit{characteristics that provide some information about the user, but lack the distinctiveness and permanence to sufficiently differentiate two individuals}'' \cite{jain04BTHI}. Figure \ref{fig:soft} illustrates examples of soft biometric attributes that have been extracted from primary biometric modalities. Examples include gender, ethnicity, age, stride length, weight, eye color, hair color, clothing, facial accessories, etc. While soft biometric traits are not discriminative enough to uniquely identify a subject, they have often been used in conjunction with primary biometric modalities to complement their performance \cite{dantcheva16Tifs}. One way to utilize soft biometric traits is by utilizing the extracted information - such as gender, ethnicity, skin color, or hair color - to reduce the search space for a given probe sample. Another commonly used technique is to extract soft biometric features and fuse them with the primary biometric trait in order to enhance identification (Figure \ref{fig:softFusion}). In some cases, soft biometric attributes can be gleaned from low-resolution biometric data \cite{singh17Ijcnn}. A review of techniques for extracting and using soft biometrics, especially in the case of face recognition, are provided in \cite{sosa18Tifs, guo18Isba, zhang15Survey, dantcheva16Tifs, reid2013Survey}. Table \ref{tab:soft} presents examples of recent papers that utilize soft biometric information for recognition. 

One of the initial approaches involving soft biometrics (referred to as `soft measure') for identification was presented by Heckathorn \textit{et al.} \cite{heckathorn01methodology}. The authors proposed using attributes such as scars, birthmarks, tattoos, eye color, ethnicity and gender, along with five biometric measures of height, forearm length and wrist width, for identifying a given subject. The model was shown to be useful in scenarios where biometric scanners are unavailable and there is a need for maintaining anonymity by eliminating the storage of biometric photographs. The model was built upon the concept of ``\textit{interchangeability of indicators}'', which states that ``\textit{indicators of low accuracy can produce, in combination, a highly accurate indicator}''.	Jain \textit{et al.} \cite{jain04BTHI} presented one of the first papers that explored the possibility of fusing soft biometric attributes with primary biometric traits for enhancing the recognition performance of an automated system. A probabilistic model based on the Bayes rule was used for combining the scores generated by the soft biometric and primary biometric systems. Experiments were performed on a fingerprint dataset of 160 subjects, with gender, ethnicity and height as soft biometrics. It was observed that the utilization of additional information enhances the recognition performance by almost 6\%. 
A similar model was later used by Guo \textit{et al.} \cite{guo10cross} for analyzing the effect of race, gender, height, and weight on cross-age face recognition. Benchmark dataset and results were also provided for the said problem.  

\begin{table*}
\centering
\label{tab:soft}
\caption{Examples of algorithms utilizing soft biometric information in conjunction with primary biometric modalities for performing recognition.}
\begin{tabular}{|c|l|l|}
\hline
\textbf{Year} & \textbf{Authors} & \textbf{Description} \\
\hline
\hline
2001 & Heckathorn \textit{et al.} \cite{heckathorn01methodology} & First study demonstrating effectiveness of soft biometrics for recognition \\
\hline
2004 & Jain \textit{et al.} \cite{jain04BTHI} & Bayes rule based model for fingerprint recognition with soft and hard biometrics \\
\hline
2004 & Zewail \textit{et al.} \cite{zewail04symp} & Utilized iris color for performing multimodal recognition of fingerprint and iris \\
\hline
2004 & Jain \textit{et al.} \cite{jain04iwba} & Utilized gender, height, and ethnicity for fingerprint and face identification \\
\hline
2006 & Ailisto \textit{et al.} \cite{Ailisto06PRL} & Incorporated weight and fat percentage for performing fingerprint recognition \\
\hline
2009 & Marcialis \textit{et al.} \cite{marcialis09group} & Proposed  \textit{minority groups} to reduce the false rejection rate using soft biometrics \\
\hline
2009 & Abreu \textit{et al.} \cite{abreu09improving} & Feature selection using soft biometrics \\
\hline
2010 & Moustakas \textit{et al.} \cite{moustakas10SPL} & User height and stride for supplementing gait recognition \\
\hline
2010 & Park and Jain \cite{park10Tifs} & Combined facial marks with an existing face recognition algorithm \\
\hline
2010 & Guo \textit{et al.} \cite{guo10cross} & Analyzed effect of race, gender, height, and weight for cross-age recognition  \\
\hline
2011 & Scheirer \textit{et al.} \cite{scheirerIjcb11} & Bayesian Attribute Networks for using descriptive attributes for face recognition \\
\hline
2011 & Abreu \textit{et al.} \cite{abreu2011enhancing} & Proposed three methods for fusing soft biometric information with primary biometrics \\
\hline
2014 & Tome \textit{et al.} \cite{tome14Tifs} & Evaluated effect of soft biometrics for recognition from a distance \\
\hline
2015 & Tome \textit{et al.} \cite{tome15Fsi} & Fusion of continuous and discrete soft biometric traits for face recognition \\
\hline
2017 & Mittal \textit{et al.} \cite{mittal17If} & Utilized soft biometrics for re-ordering the rank list of a face recognition model\\
\hline
2017 & Hu \textit{et al.} \cite{hu17Iccv} & Tensor-based fusion of face recognition features and face attribute features \\
\hline
2017 & Schumann and Stiefelhagen \cite{schumann17cvprw} & Weighted fusion of attribute prediction and face features for person re-identification \\
\hline
2018 & Kazemi \textit{et al.} \cite{kazemi18cvprw} & Attribute centered loss for CNNs: match digital faces with sketch-attribute pairs \\
\hline
2018 & Liu \textit{et al.} \cite{liu18ijcai} & Attribute guided triplet loss for heterogeneous face matching \\
\hline
\end{tabular}
\end{table*} 
 
Shortly after demonstrating the viability of fusing soft biometrics with a unimodal biometric system, Zewail \textit{et al.} \cite{zewail04symp} demonstrated the effectiveness of incorporating the iris color in a multi-modal biometric system consisting of fingerprint and iris. Fusion was performed either via a weighted average of the scores or via a Parzen Classifier. Jain \textit{et al.} \cite{jain04iwba} proposed utilizing the Bayesian model presented earlier \cite{jain04BTHI} for combining the soft biometric attributes of gender, height and ethnicity with fingerprint and face independently, as well as in a multi-modal configuration. Experimental results demonstrated the benefit of combining soft biometrics with both unimodal and multi-modal identification systems.   

In 2006, Ailisto \textit{et al.} \cite{Ailisto06PRL} analyzed the effect of including weight and fat percentage in a fingerprint recognition system. Multiple fusion techniques were explored at the decision level: AND, OR, and weighted sum. Score level fusion was also analyzed with the help of multilayer perceptrons and SVMs. Experiments across different fusion levels reiterated the benefit of using soft biometric attributes in conjunction with primary biometrics. Extending to other modalities, Moustakas \textit{et al.} \cite{moustakas10SPL} proposed utilizing the user height and stride length information for supplementing gait recognition. A probabilistic framework was used for this purpose. In 2010, Park and Jain \cite{park10Tifs} used demographic information (gender and ethnicity) and facial marks (scars, moles, and freckles) to generate a 50-bin histogram. A soft biometric matcher was then created, the score of which was fused with that of a face matcher. This combination  was observed to improve biometric performance. Scheirer \textit{et al.} \cite{scheirerIjcb11} utilized Bayesian Attribute Networks for combining multiple descriptive attributes for face identification. Descriptive attributes refer to both soft biometric traits and some non-biometric attributes as well. A noisy-OR formulation was presented that demonstrated an improvement of over 32\% when compared to a face recognition algorithm. In 2014, Tome \textit{et al.} \cite{tome14Tifs} evaluated the effect of soft biometrics on the performance of face recognition when capturing data at varying distances from the camera. A number of soft biometric attributes were considered. These attributes were grouped into three categories: \textit{global, body,} and \textit{head}. Score level fusion was then used to combine soft biometric information with face matchers. Sum rule, adaptive switch fusion rule and a weighted fusion rule were explored, where the benefit of incorporating soft biometrics was especially significant when performing recognition at larger distances. 

\begin{figure*}
\begin{center}
\includegraphics[width=7in]{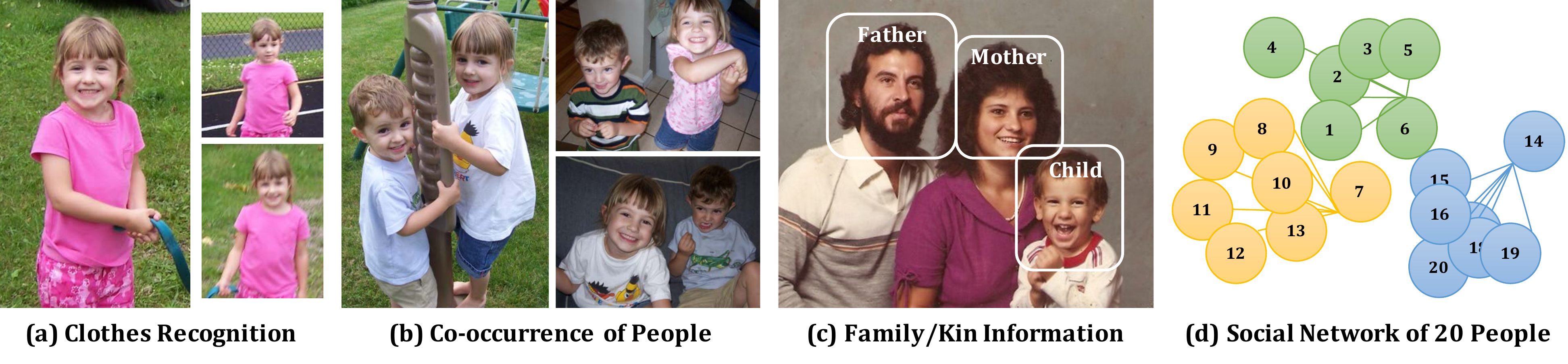}
\caption{Examples of contextual information that can be incorporated in order to enhance the biometric recognition performance. Images are taken from the {\em Images of Groups} \cite{gallagherCvpr08} and {\em Gallagher Collection Person} \cite{gallagherCvpr09} datasets.}
\label{fig:context}
\end{center}
\end{figure*}

Soft biometric traits have been used in other ways also. In 2009, Marcialis \textit{et al.} \cite{marcialis09group} demonstrated that using soft biometrics such as ethnicity and hair color with a face recognition system can help reduce the False Rejection Rate (FRR) of some users, without significantly affecting the False Acceptance Rate (FAR). A probabilistic framework was presented to predict whether an input face image belonged to a particular user, based on the extracted set of biometric features and the presence of certain soft biometric attributes. Since some soft biometric attributes (e.g., a specific hair color) are associated with only a small number of users (and not with others), it is possible to use such attributes to improve recognition accuracy.  Abreu \textit{et al.} \cite{abreu09improving} utilized soft biometric attributes for feature selection, and augmented primary biometric features with the extracted soft biometric features. Experiments in the context of signature biometrics demonstrated that utilizing soft biometric traits increases identification accuracy. Further, the authors evaluated three methods for fusion: majority-based fusion, sum-based fusion and a sensitivity-based negotiation model \cite{abreu2011enhancing}. In 2017, Mittal \textit{et al.} \cite{mittal17If} proposed using soft biometrics such as gender, ethnicity, and skin color as a way to re-order the ranked identity list generated by a face matcher. The authors demonstrated improved performance on the problem of composite sketch recognition, where the proposed technique outperformed other algorithms for the said task. In 2018, Swearingen and Ross \cite{SwearingenRoss2018} used a label propagation scheme to deduce missing soft biometric labels (viz., gender and ethnicity) by combining face data with demographic data in a graph-like structure. 

With the increased focus on deep learning techniques, some recent algorithms have incorporated soft biometric information into the deep learning pipeline. The availability of large-scale datasets with attribute information has further facilitated research in this direction \cite{liu15iccv}. Hu \textit{et al.} \cite{hu17Iccv} proposed a tensor-fusion based framework for combining face recognition and face attribute features, resulting in a Gated Two-stream Neural Network. Schumann and Stiefelhagen \cite{schumann17cvprw} proposed learning \textit{attribute-complementary} features for person re-identification. An attribute classifier is trained for different attributes such as \textit{male}, \textit{long hair}, and \textit{sunglasses}, followed by a person recognition network. Extracted attributes are provided as input to the recognition module, while learning weights for each attribute, in order to control their influence. Kazemi \textit{et al.} \cite{kazemi18cvprw} proposed an attribute-centered loss for training Deep Coupled Convolutional Neural Networks for matching digital face images against forensic sketches. Attribute information about the forensic sketch is used in a pair-wise fashion to learn a shared latent space consisting of several distinct centroids. For matching heterogeneous face images, soft biometric information has been incorporated in the triplet loss function \cite{liu18ijcai}.     

The effectiveness of combining soft biometrics with primary biometric traits in improving recognition performance can thus be observed across different studies. It is interesting to note that while initial research began in the domain of fingerprint recognition, the effect of soft biometrics is now prominent across the face, iris, and gait modalities as well. Majority of the research, however, has focused on incorporating soft biometrics in a unibiometric system. While some studies have demonstrated improvement in the case of multimodal systems also, dedicated research in this direction is necessary to yield improved performance. Most of the techniques assume prior knowledge about soft biometric attributes at the time of recognition. Developing an integrated system capable of predicting soft biometric information from the input data followed by its fusion in the biometric recognition pipeline might result in more practically deployable systems.

\subsection{Biometrics and Contextual Attributes}
Generally, context is used \textit{``to imply acceptable co-occurrence of various parts
or attributes of an object or face"} \cite{bharadwajIjcb14}. This is particularly helpful in scenarios where the identities of people in a photograph have to be deduced. Contextual attributes have been used as ancillary information along with biometric features in an attempt to aid identification performance. As shown in Figure \ref{fig:context}, different kinds of contextual information have been used for enhancing recognition performance as well as for automatically tagging images in albums or family photographs. In early work, context was established in terms of temporal, spatial, and even social information. {\em Temporal} refers to the time of capture of images, {\em spatial} corresponds to the location where the image was captured, and {\em social} refers to some information regarding the individuals present in the images or the photographer. However, the notion of context has evolved over time. Table \ref{tab:context} presents examples of techniques incorporating context in the biometric recognition pipeline.  


\begin{table*}
\centering
\caption{Examples of techniques incorporating contextual information (e.g., temporal, spatial, social) in the biometric recognition pipeline.}
\label{tab:context}
\begin{tabular}{|c|l|l|}
\hline
\textbf{Year} & \textbf{Authors} & \textbf{Description} \\
\hline
\hline
2003 & Zhang \textit{et al.} \cite{zhangMM03} & Probabilistic Bayesian framework which incorporates clothing for face tagging \\
\hline
2005 & Davis \textit{et al.} \cite{davisMM05} & Incorporated temporal, spatial, and social metadata for face recognition  \\
\hline
2006 & Song and Leung \cite{songEccv06} & Novel clothes recognition algorithm fused with face recognition using spectral clustering  \\
\hline
2007 & Anguelov \textit{et al.} \cite{anguelovCvpr07} & Markov Random Field based technique for fusing clothing and facial features  \\
\hline
2008 & Gallagher \textit{et al.} \cite{gallagherCvpr08} & Clothing segmentation algorithm using graph cuts fused with facial regions for recognition \\
\hline
2008 & Stone \textit{et el.} \cite{stoneCvprw08} & Utilized social media networks for automated tagging of images \\
\hline
2009 & Kapoor \textit{et el.} \cite{kapoorIccv09} & Incorporate logical contextual constraints into active learning for tagging photographs \\
\hline
2010 & Wang \textit{et al.} \cite{wangEccv10} & Utilize social familial context for aiding face recognition  \\
\hline
2011 & Scheirer \textit{et al.} \cite{scheirerIjcb11} & Bayesian weighting approach incorporating soft biometric traits and other attributes \\
\hline
2012 & Chen \textit{et al.} \cite{chenMM12} & Graph-based technique for predicting pair-wise relationships to improve recognition \\
\hline
2014 & Bharadwaj \textit{et al.} \cite{bharadwajIjcb14} & Social context based re-ranking algorithm using association rules \\
\hline
2014 & Hochreiter \textit{et al.} \cite{hochreiter14} & Structural Support Vector Machine incorporates album based personal and social costs \\
\hline
2015 & Bhardwaj \textit{et al.} \cite{bhardwajIcb15} & Fusion of recognition scores obtained from a social graph and face recognition algorithm \\
\hline
2016 & Li \textit{et al.} \cite{liCvpr16} & Multi-level contextual information for person, photo, and photo-group is used to aid recognition \\
\hline
2017 & Kohli \textit{et al.} \cite{kohli17Tip} & Fusion of kinship verification and face verification scores \\
\hline
2017 & Nambiar \textit{et al.} \cite{nambiar17fg} & Context-specific score-level fusion for gait recognition \\
\hline
2017 & Li \textit{et al.} \cite{li17sequential} & Person recognition in photo album using relation and scene context \\
\hline
2018 & Sivasankaran \textit{et al.} \cite{sivasankaran18Icb} & Incorporated context for continuous authentication with multiple classifiers \\
\hline
2018 & Sankaran \textit{et al.} \cite{sankaran18Icb} & Siamese architecture incorporating metadata for face recognition\\
\hline
2018 & Sultana \textit{et al.} \cite{sultana18TMan} & Fusion of face and ear biometrics with social network information \\
\hline
\end{tabular}
\end{table*}

One of the initial algorithms incorporating contextual information to aid face recognition was proposed by Zhang \textit{et al.} \cite{zhangMM03}. The authors utilized extended face region (specifically, the clothes) as ancillary information for face recognition. A semi-automated model was proposed that performs face tagging in family photos. The model presents a candidate list of potential subjects, out of which the user is asked to select the correct identity. The proposed framework is built over a probabilistic Bayesian framework which works with both facial and contextual features. Davis \textit{et al.} \cite{davisMM05} also proposed a semi-automated model for performing face tagging in photographs. The authors incorporated temporal, spatial, as well as social meta-data to aid in face recognition.  Here, temporal refers to the exact time the photo was taken as per the cellular network; spatial refers to the Cell ID from the cellular network and location from Bluetooth-connected GPS receivers; and social refers to the identity of the photographer, the sender and recipients (if any) of the photo, and those who were co-present when the photo was taken (sensed via Bluetooth MAC addresses mapped to usernames). A specific logger was designed and installed on cell-phones to track the aforementioned meta-data. Sparse-Factor Analysis (SFA) was used to perform face recognition using a combination of facial features and contextual meta-data. Experimental evaluation conveyed that utilizing contextual information improves the performance of face recognition compared to using either of the information independently. 

In 2006, Song and Leung \cite{songEccv06} proposed a model which fused clothing information with face recognition results in order to perform improved person identification. A novel `clothes recognition' algorithm was proposed, the results of which were integrated into a spectral clustering algorithm to perform person recognition. Logic-based constraints, such as requiring different individuals in a photograph to correspond to different identities, were also enforced by the clustering algorithm. The authors show that the performance of the clustering algorithm for face-based recognition improves with clothing information is provided and logic-based constraints are imposed. 

In 2007, Anguelov \textit{et al.} \cite{anguelovCvpr07} proposed a Markov Random Field (MRF) based technique which combines clothing and facial features in order to generate a probabilistic model for predicting the identity corresponding to a given face image. Temporal context in terms of time stamps were used to create \textit{events} based on the clothing of individuals. Different features were used for encoding the color and texture of the clothes, and the Loopy Belief Propagation (LBP) method was used for performing MRF inference. Gallagher \textit{et al.} \cite{gallagherCvpr08} proposed a clothing segmentation algorithm based on graph cuts. Features were extracted from the facial and clothing regions, and a probabilistic model was trained to perform person recognition. The authors proposed using the algorithm in the case of consumer image collections, where the number of individuals in an image are known and some individuals have already been labeled by the user. Thus, the task of the model is to detect and label the remaining individuals.

Most of the work until 2008 focused on utilizing spatial and temporal information - such as timestamps of images, geographical location, co-occurrence of individuals and personal clothing - to incorporate contextual information for aiding face recognition in group photo collections. Motivated by the large-scale availability of meta-data on social media sites such as FaceBook, Stone \textit{et al.} \cite{stoneCvprw08} proposed a technique to utilize contextual information for complementing face recognition algorithms and automatically tagging face images. The authors collected images and meta-data from a fixed set of FaceBook users. The tagged images were then used to train a Conditional Random Field (CRF) algorithm built upon pairwise links between faces observed in photographs. The authors observed improved face recognition performance when incorporating contextual information in the proposed model. 

Kapoor \textit{et al.} \cite{kapoorIccv09} proposed the incorporation of logical contextual constraints into the paradigm of active learning to tag group photographs. The framework was presented for performing face tagging on personal photo and video collections using \textit{match} and \textit{non-match} constraints based on prior information. Further, Wang \textit{et al.} \cite{wangEccv10} presented a unique formulation for incorporating social familial context into a face recognition pipeline. The authors consider the scenario where weak labels are provided for a given image, and which need to be assigned to each face present in the image. Familial social relationships, such as ``mother-child'' or ``siblings'', are used to infer the relative positioning of the face images associated with their corresponding labels. A graphical model is trained for each individual that utilizes facial features as well as features that reflect social relationships. Experimental results on datasets containing consumer images convey the efficacy of the proposed method.

\begin{table*}
\centering
\label{tab:continuous}
\caption{A brief summary of techniques used for continuous authentication in a multimodal setting.}
\begin{tabular}{|c|l|l|}
\hline
\textbf{Year} & \textbf{Authors} & \textbf{Description} \\
\hline
\hline
2003 & Altinok and Turk \cite{altinok03} & Bayes classifier based technique for temporal integration of multiple modalities across time \\
\hline
2007 & Sim \textit{et al.} \cite{sim07Pami} & Holistic fusion method built over Hidden Markov Models  \\
\hline
2008 & Azzini \textit{et al.} \cite{azzini08fuzzy} & Fuzzy controller based model for dynamically switching between modalities \\
\hline
2009 & Kwang \textit{et al.} \cite{kwang09Icb} & Study on usability of a continuous authentication based system over 58 participants \\
\hline
2010 & Niinuma \textit{et al.} \cite{niinuma10Tifs} & Combines soft biometric traits of face and clothing color with PCA based face recognition  \\
\hline
2010 & Shi \textit{et al.} \cite{shi10Security} & Learns user profile based on mobile phone usage habits using Gaussian Mixture Models \\
\hline
2013 & Frank \textit{et al.} \cite{frank13Tifs} & Proposes a set of 30 behavioral touch features for smart-phone authentication  \\
\hline
2016 & Sitov{\'a} \textit{et al.} \cite{sitova16Tifs} & Hand Movement, Orientation, and Grasp (HMOG) for smartphones \\
\hline
2017 & Peng \textit{et al.} \cite{peng17Thms} & \textit{GlassGuard}: Continuous authentication system for Google Glass using touch and voice features \\
\hline
2017 & Fenu \textit{et al.} \cite{fenu17Prl} & Multimodal fusion technique for face, voice, touch, mouse, and keystroke based features \\
\hline
2018 & Kumar \textit{et al.} \cite{kumar18Isba} & Score-level fusion of one-class classifiers \\
\hline
2018 & Shen \textit{et al.} \cite{shen18Tifs} & Feature-level combination of multi-motion sensor behavior \\
\hline
\end{tabular}
\end{table*}

Scheirer \textit{et al.} \cite{scheirerIjcb11} proposed incorporating soft biometric traits, descriptive attributes, and contextual information for face recognition. A novel Bayesian weighting approach was proposed which provides weights to the scores obtained for all the samples in the database based on the attribute network of the gallery images and attributes/context based features extracted from the probe. Chen \textit{et al.} \cite{chenMM12} proposed a graph-based technique for predicting the pair-wise relationship between two faces in a group photograph. The proposed model generates graphs and sub-graphs, in order to understand the social relationships between people, from a given set of group photographs. The authors also propose a Bag of Face subGraph (BoFG) which is based on the co-occurrence of individuals in different photographs. For a given test image, the BoFG is calculated and classification is performed based on a Naive Bayes classifier. The authors present improved performance, compared to other techniques utilizing only descriptive visual features for performing the same task. 
 
Bharadwaj \textit{et al.} \cite{bharadwajIjcb14} proposed a social context based re-ranking algorithm for improving the classification performance of any classifier by incorporating context based rules. Association rule mining is used for inferring associations between individuals in group photographs. Multiple rules are generated and utilized to obtain context based scores. At the time of testing, these scores are combined with the normalized scores obtained from the classifier, in order to re-rank the results provided by the classifier. Hochreiter \textit{et al.} \cite{hochreiter14} also proposed a technique to incorporate album based costs in a recognition framework. Two types of costs, personal and social, are included in the optimization of a structural SVM, in order to include contextual information obtained from albums of photographs. 
 
Another algorithm for updating the rankings obtained from an existing face recognition system was proposed by Bhardwaj \textit{et al.} \cite{bhardwajIcb15}. The proposed technique utilizes a social graph (created from training images), where each node is treated as a subject, in order to learn the contextual information between the subjects. For a given group photo, the face recognition scores returned from a traditional face recognition system are combined with those obtained from the social graph, in order to perform context-aided face recognition. Recently, Li \textit{et al.} \cite{liCvpr16} proposed a novel framework for utilizing multi-level contextual information at the person, photo, and photo group levels. At the person level, the algorithm utilizes contextual information related to clothes and body appearance, while in photographs of groups, a joint distribution of identities as well as meta-data is used to guide the recognition task. The authors present a framework consisting of SVMs and Conditional Random Fields to incorporate the aforementioned levels of contextual information in the recognition pipeline. 

In 2017, Kohli \textit{et al.} \cite{kohli17Tip} proposed incorporating kinship verification scores as contextual information in the face verification pipeline. A deep learning based framework was used for kinship verification, followed by a score-level fusion with face verification via the product of likelihood ratio and SVM-based approaches. Recently, context information has been incorporated into a classifier ensemble for person re-identification or continuous authentication \cite{nambiar17fg,sivasankaran18Icb}. Sankaran \textit{et al.} \cite{sankaran18Icb} proposed a Siamese convolutional neural network which utilized meta-data of face images such as \textit{yaw}, \textit{pitch}, and \textit{face size} to enhance face recognition. Sultana \textit{et al.} \cite{sultana18TMan} proposed incorporating social behavioral information extracted from online social networks in a multi-modal system based on face and ear recognition. Scores of different modalities were fused at the score-level in order to obtain the final decision. 

It is interesting to note the progression of what constitutes as contextual information across time. While initial research began with incorporating clothing related information in the recognition pipeline, researchers have now started utilizing social network graphs as well. Temporal information, such as the time of capture, remains an important feature for categorization of photographs into \textit{events} in the case of tagging multiple images. Logical constraints, such as ensuring that different faces in a photograph belong to different individuals, are also often utilized by such algorithms. With the advent of social media, and easy availability of related meta-data, a majority of recent techniques have focused primarily on social networks to aid in the recognition process. Such algorithms implicitly assume an active social media presence, thereby restricting their usability. A combination of contextual information derived from social media and traditional approaches could further enhance recognition performance and improve response time for a query.

\subsection{Continuous Authentication}
\begin{figure}
\begin{center}
\includegraphics[width=3.2in]{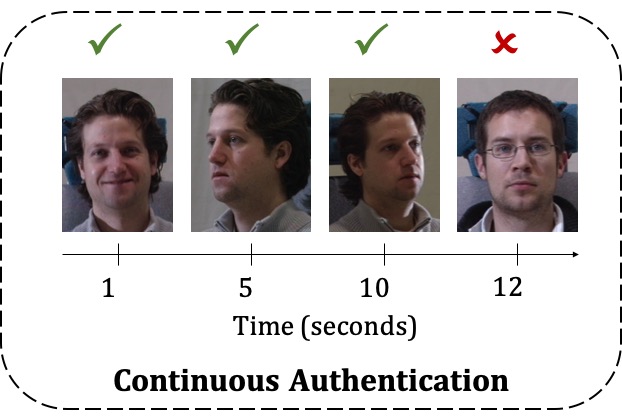}
\caption{In a continuous authentication system, the identity of the user is verified at regular intervals. In this illustration, the user is rejected access at t=12 seconds since the system was (correctly) unable to confirm his identity.}
\label{fig:continuous}
\end{center}
\end{figure}

In some high security applications, it is necessary for access to be restricted to specific individuals. In such scenarios, there is often a need for authenticating the identity of an individual multiple times. For instance, when accessing confidential data over a length of time using a device, an individual may have to be \textit{continuously} authenticated to ensure that an unauthorized adversary does not view the data during the transaction. It is, therefore, not sufficient for the individual's identity to be authenticated only at the beginning of the session - authentication has to occur at periodic intervals during the entire session (Figure \ref{fig:continuous}). Depending upon the task at hand, it might be difficult to obtain continuous data pertaining to a single biometric modality. Therefore, multiple cues are needed to facilitate user authentication in such scenarios. One of the initial papers on continuous authentication using multimodal biometrics was by Altinok and Turk \cite{altinok03}. The authors proposed a temporal integration technique for performing continuous authentication using multiple biometric cues - face, voice and fingerprint. A Bayes classifier was used for combining the normalized match scores across the 3 channels, i.e., the 3 biometric cues. Then a temporal integration method was used to generate an expected score distribution and an estimated uncertainty of the distribution. Estimates were calculated as a function of the previous observation and the current time, in order to encode the temporal dependence between observations as well. 

In 2007, Sim \textit{et al.} \cite{sim07Pami} developed a \textit{holistic fusion} method built over Hidden Markov Models for integrating evidence from the face and fingerprint modalities over time. Another interesting technique for handling multiple modalities across time was presented by Azzini \textit{et al.} \cite{azzini08fuzzy}. The authors proposed a fuzzy controller based model which performed decision level fusion of multiple modalities. The model was built over a \textit{trust} parameter based on which the fuzzy controller decided whether to perform authentication using a single modality or via fusion of multiple modalities. When the \textit{trust} value goes beyond a pre-defined threshold for all scenarios, the user is logged off the system. Kwang \textit{et al.} \cite{kwang09Icb} performed a study on the usability of continuous authentication systems in real life. A study was performed on 58 participants wherein they were required to perform certain tasks on a Windows machine equipped with a Continuous Biometrics Authentication System (CBAS). The authors concluded in favor of using a CBAS despite having substantial system overhead.  

Most of the research until 2010 focused on fusing primary biometric traits, such as fingerprint and face, for performing continuous authentication. Depending upon the system at hand and acquisition environment, obtaining a good quality face or fingerprint template \textit{continuously} might not be a viable assumption. One can expect pose and illumination variations, incomplete or no capture, or even forced repeated co-operation from users. Inspired by these observations, Niinuma \textit{et al.} \cite{niinuma10Tifs} proposed utilizing soft biometric traits for performing continuous authentication. The proposed framework combines face color and clothing color with PCA based face recognition. Continuous authentication is performed using the soft biometric traits, and face recognition is used for template enrollment and re-authentication. Score level fusion is utilized for continuous authentication, which is governed by a fixed threshold; a re-login is requested when authentication fails. Shi \textit{et al.} \cite{shi10Security} proposed a framework for learning user profiles based on their usage habits on mobile phones. Information such as call logs, SMS logs, browsing habits and GPS location are used to generate features for learning each user's profile. In real time, each event generates a score value, based on which the identity of the user is either confirmed or refuted. The authors also evaluate the robustness of their method to different types of adversarial attacks. 


In 2013, Frank \textit{et al.} \cite{frank13Tifs} proposed using a set of 30 behavioral biometric features for performing continuous authentication on smartphones. User interaction with the touch screen of their smartphone is analyzed to develop a set of different strokes that are classified using the KNN and SVM classifiers. A combination of hand movement, orientation, and grasp (HMOG) features have also been used to continuously authenticate a user on a smartphone \cite{sitova16Tifs}. Data collected from the accelerometer, gyroscope, and magnetometer is combined to perform unobtrusive authentication. Patel \textit{et al.} \cite{patel16Spm} presented a survey on continuous authentication focusing on both unimodal and multimodal techniques. The authors discuss the progress in the field along with existing methods and related challenges. With the development in technology, the domain of continuous authentication has further expanded to incorporate wearable devices such as Google Glass as well. Peng \textit{et al.} \cite{peng17Thms} proposed \textit{GlassGuard} which fuses decision scores obtained from touch gestures and voice commands in a probabilistic manner to authenticate the user's identity. Researchers have also worked on continuous authentication for e-learning platforms \cite{fenu17Prl}, where a multimodal fusion technique utilizing face, voice, touch, mouse, and keystroke is proposed. Independent verification scores are calculated for each modality which are then combined via a score fusion mechanism. In order to reduce the computational cost, verification is performed at predefined time intervals. 

Table \ref{tab:continuous} presents a brief summary of algorithms proposed for continuous authentication. It can be observed that the field of continuous authentication has evolved tremendously over the past decade. Researchers have attempted to address the problem in desktops, laptops, smart phones, and even wearable devices. While there exists several sophisticated algorithms in the literature, a major challenge is the lack of publicly available large datasets for the given problem. Computational efficiency in continuous authentication further remains as one of the major challenges requiring dedicated attention.

\begin{figure*}[t]
\begin{center}
\includegraphics[width=6.6in]{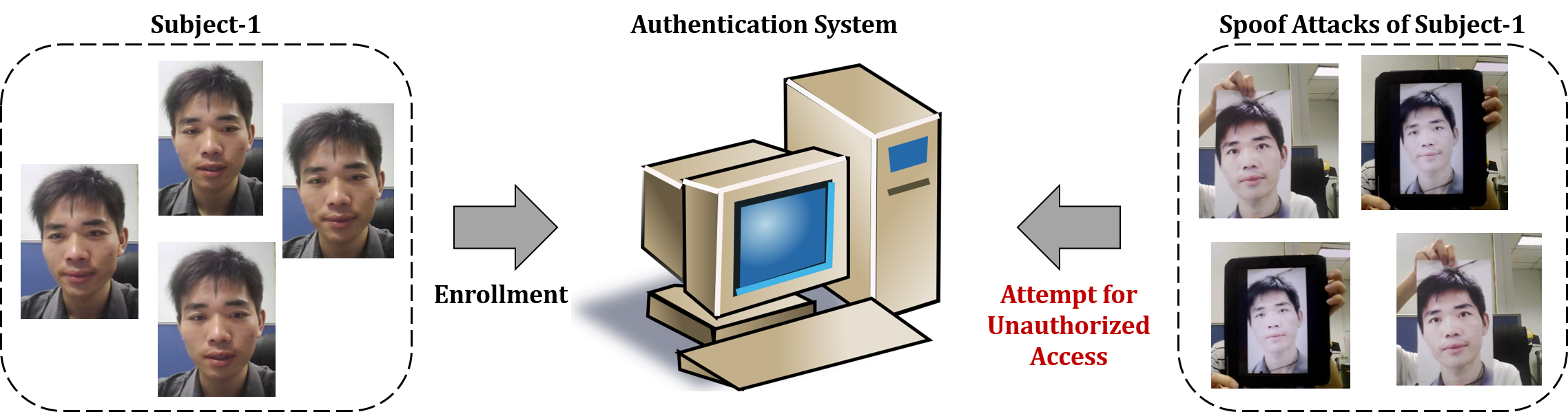}
\caption{Presentation attacks on a face recognition system where an adversary attempts to spoof the identity of Subject-1. Face images have been taken from the CASIA-FASD database \cite{casiaSpoofDb}.}
\label{fig:spoof}
\end{center}
\end{figure*}

\section{Biometric Fusion and Presentation Attack Detection}
\label{sec:spoof}

A biometric system is vulnerable to a number of attacks~\cite{Ratha2001}. One such attack is referred to as a {\em presentation attack} where an adversary presents a fake or altered biometric trait to the sensor with the intention of spoofing someone else's trait; or creating a new virtual identity based on the presented trait; or obfuscating their own trait. Detecting such attacks is essential in improving the security and integrity of the biometric system. In earlier literature, {\em presentation attack} was synonymous with {\em spoof attack} and, therefore, the terms {\em spoof detection} and {\em anti-spoofing} have been used in connection with presentation attack detection (PAD).   


\begin{table*}[t]
\centering
\label{tab:spoof}
\caption{A summary of some techniques that use fusion for either spoof (i.e., presentation attack) detection or for combining the anti-spoofing module of a biometric system with the biometric matcher itself.}
\begin{tabular}{|c|l|l|}
\hline
\textbf{Year} & \textbf{Authors} & \textbf{Description} \\
\hline
\hline
2012 & Marasco \textit{et al.} \cite{marasco12Btas} & Different frameworks for integrating a spoof detection module with a recognition system \\
\hline
2015 & Wen \textit{et al.} \cite{wen15Tifs} & Ensemble of SVMs on reflection, blurriness, chromatic moment, and color diversity \\
\hline
2015 & Raghavendra \textit{et al.} \cite{raghavendra15Tip} & Feature level concatenation with Light Field Camera based features \\
\hline
2015 & Arashloo \textit{et al.} \cite{arashloo15Tifs} &  Fused MBSIF-TOP and MLPQ-TOP using SR-KDA \\
\hline
2016 & Ding \textit{et al.} \cite{DingBBN_ICB2016} & Bayesian Belief Networks for fusing match scores with liveness scores \\
\hline
2016 & Boulkenafet \textit{et al.} \cite{boulkenafet16Tifs} & CoALBP and LPQ features in HSV and YCbCr colour space\\
\hline
2016 & Patel \textit{et al.} \cite{patel16Tifs} & Concatenation of color moments and LBP features\\
\hline
2016 & Siddiqui \textit{et al.} \cite{siddiqui16Icpr} & Inter-feature and intra-feature score-level fusion of multi-scale LBP and HOOF features \\
\hline
2016 & Ding and Ross~\cite{DingOCSVM2016} & Fusion of multiple one-class SVMs to improve generalizability of a fingerprint spoof detector \\
\hline
2017 & Toosi \textit{et al.} \cite{toosi17Access} & Comparative study of different fusion techniques on ten fingerprint features \\
\hline
2017 & Korshunov and Marcel \cite{korshunov17Stsp} & Studies impact of score fusion on presentation attack detection for voice \\
\hline
2018 & Komeili \textit{et al.} \cite{hatzinakos18Tifs} & Fusion of ECG recognition and fingerprint spoof detection \\
\hline
2018 & Yadav \textit{et al.} \cite{yadav18Cvprw} & Fusion of (VGG features+PCA) with (RDWT+Haralick) features and neural network \\
\hline
2018 & Sajjad \textit{et al.} \cite{sajjad18Prl} & Two-tier authentication system for recognition and spoof detection \\
\hline
2018 & Chugh \textit{et al.} \cite{chugh18Tifs} & CNN based spoof detection on fingerprint patches \\
\hline
\end{tabular}
\end{table*} 

The task of spoof detection has also been viewed as \textit{liveness} detection, especially in the initial research revolving around fingerprint recognition \cite{ghiani17review}. Liveness detection involves predicting whether a given sample is \textit{live/bonafide} or not, that is, whether the input is captured from a human being or is a synthetically generated artifact. Liveness or spoof detection modules can be integrated into a biometric recognition pipeline in order to create systems robust to attacks \cite{marasco2011increase,chingovska2013anti,sajjad18Prl}. Marasco \textit{et al.} \cite{marasco12Btas} presented different frameworks for integrating a spoof detector with a fingerprint recognition module. The authors evaluate different techniques including sequential methods, classifier-based fusion, and a Bayesian Belief Network (BBN) which explicitly models the relationship between the spoof detection scores and biometric match scores. In~\cite{DingBBN_ICB2016}, Ding and Ross explored multiple BBN architectures for modeling the influence of liveness scores on match scores (and vice-versa) and used these architectures for fusing the two scores. 

In the literature, the task of spoof detection has generally been handled independently for different biometric modalities. Researchers have focused on analyzing the effect of different \textit{attacks} on biometric systems for various modalities including face, fingerprint, iris, vein-pattern, hand geometry, and speech. Recent surveys on anti-spoofing algorithms for these modalities can be found in  \cite{faceSpoofSurvey, Marasco14, presentationAttackSurvey, reviewIrisSpoof, speakerSpoof}. A major section of research utilizes a combination of texture or quality based features for the given task. Most of these techniques involve feature level fusion of different descriptors, where features are first concatenated and then input to a classifier (often an SVM) for spoof detection \cite{maata11Ijcb,grag13Workshop,bharadwaj13Cvprw,agarwal16Btas,galbally14Icpr,patel15Icb}. 

Wen \textit{et al.} \cite{wen15Tifs} proposed using Image Distortion Analysis (IDA) for identifying spoofed face images. An ensemble of SVMs is developed based on four features: specular reflection, blurriness, chromatic moment, and color diversity. The authors argue that different spoof attacks might be easily identified by different features and, therefore, learned a separate SVM for each feature. Score level fusion schemes using the min-rule and the sum-rule were used for taking the final decision. Raghavendra \textit{et al.} \cite{raghavendra15Tip} proposed using Light Field Camera (LFC) for performing facial spoof detection. As opposed to regular cameras, LFCs can be used to render an image with variations in focus and depth. This characteristic enabled the authors to observe distinct differences between real and spoofed images. Specifically, for presentation attacks, where the input sensor is presented with a print-out of another person's biometric sample, feature level concatenation of estimated variations in focus resulted in improved performance. Arashloo \textit{et al.} \cite{arashloo15Tifs} proposed fusing Multiscale Binarized Statistical Image Features on Three Orthogonal Planes (MBSIF-TOP) and Multiscale Local Phase Quantization on Three Orthogonal Planes (MLPQ-TOP) for performing spoof detection. Fusion was performed by a kernel fusion approach, termed as Spectral Regression Kernel Discriminant Analysis (SR-KDA).    

In 2016, Boulkenafet \textit{et al.} \cite{boulkenafet16Tifs} proposed using local texture features of both the luminance and chrominance channels for performing facial spoof detection. Features extracted using Co-Occurrence of Adjacent Local Binary Patterns (CoALBP) and Local Phase Quantization (LPQ) were concatenated in the HSV and YCbCr color spaces. This was followed by classification using an SVM. The proposed technique achieved state-of-the-art results on three datasets, thereby demonstrating the benefit of combining texture features from different color spaces. Patel \textit{et al.} \cite{patel16Tifs} proposed concatenating color moments and Local Binary Patterns (LBP) from a given face image for performing spoof detection. An input RGB image was converted into the HSV space for calculating color moments, and the concatenated feature vector was input to an SVM for classification. Siddiqui \textit{et al.} \cite{siddiqui16Icpr} proposed a multi-feature face spoof detection algorithm consisting of a multi-scale configuration of LBP and Histogram of Oriented Optical Flow (HOOF) features classified using an SVM. Experiments were performed using spoofed and bonafide video samples, wherein intra-feature and inter-feature fusion was performed at the score level. In \cite{liudepthrppg2018}, the authors designed a novel deep learning architecture that fused a CNN with RNN in order to extract pseudo-depth images and a remote photoplethysmography (RPPG) signal from an input face video. The extracted information were then fused for face anti-spoofing.

Recently, Toosi \textit{et al.} \cite{toosi17Access} presented a comparative study with ten feature descriptors for the task of fingerprint spoof detection. The authors experimented with different fusion strategies to achieve improved performance. The authors also proposed \textit{SpiderNet}, a two-stage deep learning architecture to learn independent and combined features for different feature inputs in order to  identify spoofed images. Korshunov and Marcel \cite{korshunov17Stsp} studied the impact of score fusion for performing presentation attack detection in case of voice biometrics. The authors used eight state-of-the-art algorithms to understand the effect of mean, logistic regression, and polynomial logistic regression fusion methods. The authors also provided open source implementations of the detection algorithms, fusion techniques, and evaluation framework. A novel framework for fusing electrocardiogram (ECG) recognition with fingerprint spoof detection was proposed by Komeili \textit{et al.} \cite{hatzinakos18Tifs}. Two classifiers - one for ECG verification and the other for fingerprint spoof detection - were trained independently and score-level fusion was performed using weighted sum, product, and maximum fusion rules. Yadav \textit{et al.} \cite{yadav18Cvprw} presented a framework for iris spoof detection, where features from a deep learning algorithm, VGG, were fused with texture based RDWT + Haralick features. The features were concatenated and input to a neural network for performing iris spoof detection. Deep learning based CNNs have also been shown to perform well for fingerprint spoof detection \cite{chugh18Tifs}. Here, minutiae detection is performed on an input fingerprint image, followed by patch generation and alignment. A CNN architecture is used to generate the liveness score for each patch, followed by score-level fusion. Ding and Ross~\cite{DingOCSVM2016} proposed the use of an ensemble of one-class classifiers in order to handle the problem of limited spoof samples during training as well as address the need to develop methods for detecting previously unseen spoofs in the context of fingerprints. Each one-class classifier was based on a simple texture descriptor and was trained predominantly on bonafide fingerprint samples. Fusion of these multiple classifiers was observed to increase the robustness of the spoof detector to novel spoof fabrication materials.   

It is interesting to observe that while most of the techniques do not utilize modality-specific information for performing spoof detection, none of the papers have demonstrated results across different biometric modalities (Table \ref{tab:spoof}). Most algorithms are evaluated on controlled data collected in a laboratory environment, which often does not simulate the real world well. It is thus essential to develop datasets that better imitate real world scenarios in order to develop robust algorithms capable of improving state-of-the-art performance and demonstrating better generalization abilities. The Liveness Detection Competition Series\footnote{http://livdet.org/competitions.php} corresponds to a series of spoof detection challenges conducted regularly since 2009. The competition series aims at evaluating and benchmarking anti-spoofing algorithms for the tasks of iris and fingerprint spoof detection. With the development of multi-modal systems, algorithms must also be developed to handle spoof detection for multi-modal systems, especially when it might be easier to spoof one modality compared to the other (see~\cite{marasco2011increase,RodriguesMultimodal2010}).

Recently, the related area of adversarial detection has garnered substantial attention, especially in the domain of Deep Learning \cite{akhtar2018threat,biggio2015adversarial}. It has been shown that adversarial samples can be created by adding small perceptible or imperceptible perturbations to the input images, which can then be used to \textit{fool} a recognition system \cite{moosavi2017universal,moosavi2016deepfool}. The presence of an adversarial detection module often results in a more robust recognition system that is reasonably immune to such adversarial attacks \cite{agarwal18Btas,goswami19Ijcv}. Most of the research in the area of adversarial detection utilizing fusion has focused on the intermediate representations obtained from the learned networks. Li and Li \cite{li17Iccv} proposed using the convolutional filter outputs of a CNN model for detecting adversarial samples. Different filter outputs are used to compute statistics for a given input which are then provided to a classifier cascade for adversarial detection. Product rule fusion is applied on the scores returned by the classifiers. Goswami \textit{et al.} \cite{goswami18AAAI,goswami19Ijcv} proposed learning the difference between the mean unperturbed features and representations extracted from the adversarial samples. Features are extracted from multiple intermediate layers of a CNN model, followed by a SVM model for adversarial detection. The authors proposed \textit{selective dropout} for handling adversarial samples by mitigating the effect of the adversary. Tao \textit{et al.} \cite{tao2018attacks} proposed detecting adversarial face samples by combining attribute information in a traditional face recognition system. As can be observed, limited research has focused on utilizing information fusion techniques for adversarial detection. As described earlier, multiple details can be extracted from a biometric sample, such as soft biometric attributes or quality scores, thereby rendering it rich in information. Thus, the inclusion of such fusion methods in the adversarial detection module could result in enhanced performance, resulting in more robust recognition systems. 

\begin{table*}
\centering
\label{tab:crypto}
\caption{A few examples of techniques that have been used for multibiometric cryptosystems.}
\begin{tabular}{|c|l|l|}
\hline
\textbf{Year} & \textbf{Authors} & \textbf{Description} \\
\hline
\hline
2007 & Sutcu \textit{et al.} \cite{sutcuCvpr07} & \textit{Secure Sketch} construct for protecting face and fingerprint templates \\
\hline
2008 & Nandakumar and Jain \cite{nandakumarBtas08} & Fuzzy vault framework for securing multibiometric templates  \\
\hline
2008 & Camlikaya \textit{et al.} \cite{camlikayaSpie08} &  Encodes minutiae features (fingerprint) within a voice feature vector \\
\hline
2009 & Fu \textit{et al.} \cite{fuTifs09} &  Models for performing fusion at the cryptographic level \\
\hline
2012 & Nagar \textit{et al.} \cite{nagarTifs12} & Fusion of fuzzy vault and fuzzy commitment to generate a single secure sketch   \\
\hline
2012 & Rathgeb \textit{et al.} \cite{rathgebBook12} & Comprehensive survey of techniques used for securing biometric templates \\
\hline
2014 & Rathgeb and Busch \cite{rathgebCs14} & Bloom filter based technique for fusing multiple features \\
\hline
2014 & Chin \textit{et al.} \cite{chin14If} & Random tiling and equi-probable $2^{N}$ discretization scheme for fingerprint and palmprint \\
\hline
2015 & Li \textit{et al.} \cite{liTifs15} & Proposed technique performs security analysis on multibiometric cryptosystems \\
\hline
2016 & Kumar and Kumar \cite{kumarAccess16} & Combination of BHC encoding and hash code computation followed by cell array storage \\
\hline
\end{tabular}
\end{table*}

\begin{figure}
\begin{center}
\includegraphics[width=2.3in]{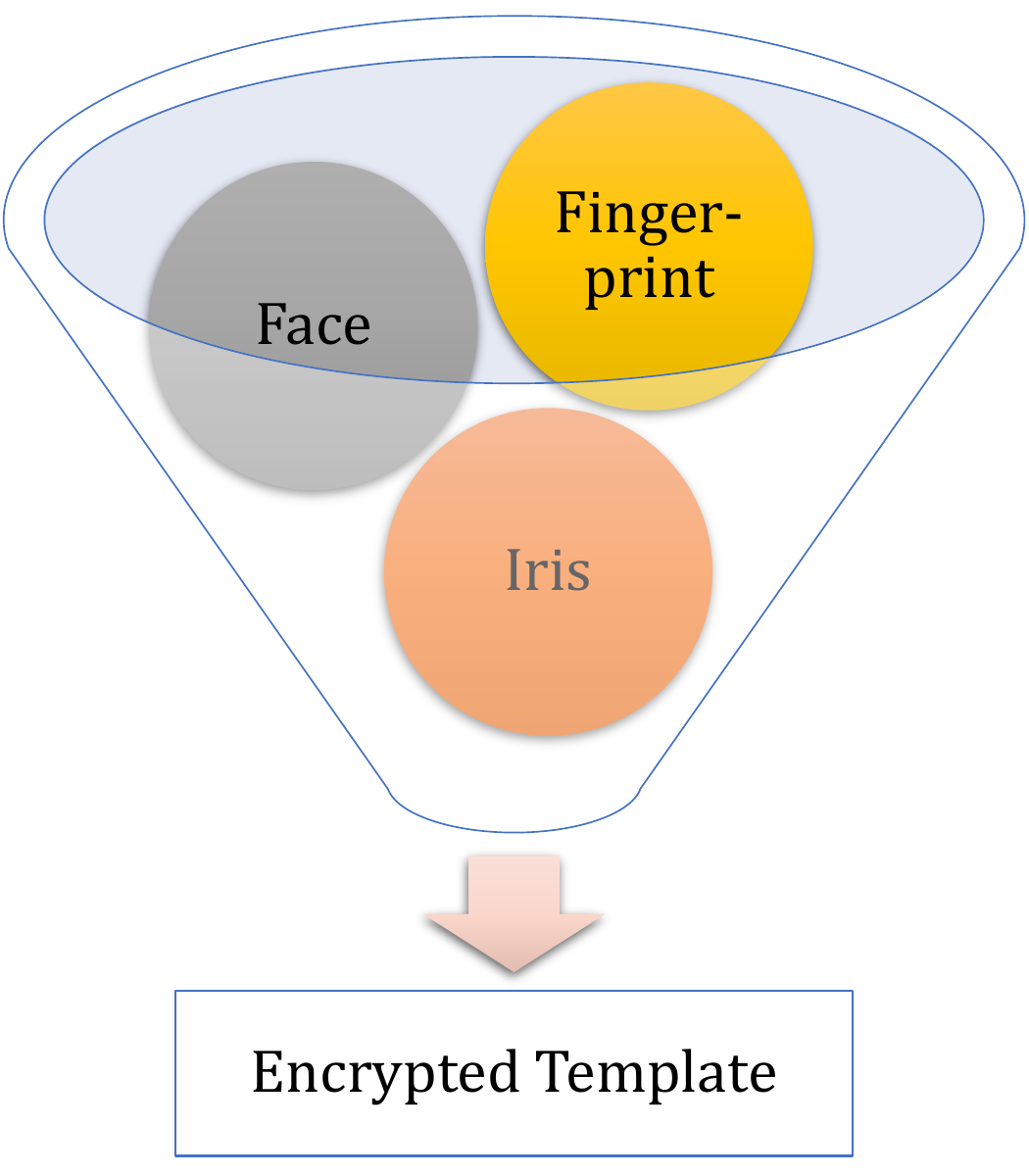}
\caption{Multibiometric cryptosystems utilize multiple biometric modalities and perform encryption in order to secure the biometric data of a subject.}
\label{fig:crypto}
\end{center}
\end{figure}

\section{Multibiometric Cryptosystems} 
\label{sec:crypto}
Data used by a biometric system can be encrypted using strong cryptographic techniques in order to secure them against external attacks. In addition, biometric matching has to be performed in the encrypted domain to obviate the need to decrypt the data, thereby preventing an adversary from viewing the original data at any time. In multibiometric systems, where data from multiple biometric sources are available, each data piece has to be encrypted. Such systems consisting of multiple biometric data sources, along with some cryptographic technique for securing the data, are termed as Multibiometric Cryptosystems. Table \ref{tab:crypto} presents a brief summary of techniques proposed for multibiometric cryptosystems. Rathgeb and Busch \cite{rathgebBook12} present a comprehensive review of the work done in the field of securing biometric templates, for both multibiometric as well as unibiometric systems. In their work, the concepts of securing a biometric template as well as a multibiometric template are explained in detail, along with a discussion of a theoretical framework for multibiometric cryptosystems.   

In order to secure multibiometric systems, Sutcu \textit{et al.} \cite{sutcuCvpr07} proposed the fusion of face and fingerprint features, followed by a \textit{secure sketch} construct for securing the fused samples, and making it difficult to reconstruct the original samples from the encrypted features. Minutiae based features were extracted from fingerprints, and Singular Value Decomposition (SVD) based features were used for the face modality. Nandakumar and Jain \cite{nandakumarBtas08} proposed a fuzzy vault framework for securing multibiometric templates consisting of fingerprint and iris. Experiments were performed using multiple impressions of the same biometric modality (fingerprint), multiple instances of a biometric modality (two index fingers), and data from multiple biometric modalities (fingerprint and iris). In the proposed technique, all features were represented as elements of a Galois Field, GF. Separate techniques were presented for feature extraction from fingerprint and iris, which were then fused at the feature level, and secured via a fuzzy vault technique. Camlikaya \textit{et al.} \cite{camlikayaSpie08} proposed a template fusion technique for the fingerprint and voice modalities. The proposed algorithm strengthened the security of the multibiometric system by encoding the minutiae features obtained from the fingerprints within the voice feature vector. The authors also motivated the chosen modalities by emphasizing the desirable cancelable property of spoken words being used as a password.    

In 2009, Fu \textit{et al.} \cite{fuTifs09} proposed several multibiometric cryptosystem models. One model for performing fusion at the biometric level was proposed, while three models were presented for performing fusion at the cryptographic level. While no experimental evaluation was performed for the proposed models, however, an in-depth analysis of the algorithms, comparisons, and discussions were presented by the authors. Nagar \textit{et al.} \cite{nagarTifs12} proposed a feature-level fusion based multibiometric cryptosystem for performing recognition using features from multiple biometric modalities. A single secure sketch was generated from multiple features (of different modalities) based on two biometric cryptosystems -  fuzzy vault and fuzzy commitment. A detailed experimental analysis was conducted on datasets pertaining to three modalities: face, iris, and fingerprints. 

Rathgeb and Busch \cite{rathgebCs14} proposed a Bloom filter based transformation technique for performing iris identification. The algorithm fused the features obtained from the left and right irides and obscured the information present in each instance independently, thus securing it against external attacks. Li \textit{et al.} \cite{liTifs15} proposed a new method for performing a security analysis on multibiometric cryptosystems, based on a combination of principles from information-theory and computational security. The authors also proposed a decision-level fusion based multibiometric cryptosystem for performing fingerprint recognition. A two stage encryption was performed on the extracted features, followed by decision-level fusion to obtain the identity of the given sample. Kumar and Kumar \cite{kumarAccess16} proposed a cell array based multibiometric cryptosystem. Bose Chaudhuri Hocquenghem (BCH) encoding and hash code computation was performed on the biometric modalities. The data was stored in the form of two cell arrays such that the hash code is distributed in the first cell array and the key is scattered across the second. Furthermore, two models were proposed, one for decision-level fusion and another for feature-level fusion. Experimental analysis depicted superiority of decision-level fusion over feature-level fusion for the proposed multibiometric cryptosystem. 

We observe that the security aspect of multibiometric systems has garnered dedicated attention. Initially, researchers applied the techniques prevalent in unibiometric systems on the fused feature vectors of different modalities. Techniques such as generation of secure sketches and fuzzy vault constructs have now been well explored. Several novel techniques have also been proposed with the aim of being more effective in terms of security and computation. In order to increase the robustness of such systems and enhance their real world applicability, research has also focused on proposing new metrics for the evaluation of the models. It is interesting to note that most of the techniques proposed in the literature for multibiometric security have focused on hand-crafted features, with a limited focus on representation learning based algorithms. This is bound to change, given the increasing interest in utilizing deep neural networks for addressing the problem of multibiometric security. 

\section{Research Challenges and Future Directions}
\label{sec:future}
Biometric fusion has witnessed significant advancements over the past two decades in terms of algorithm development, sources of information being fused, application domains and operational data collected. The literature review in Sections \ref{sec:multi}-\ref{sec:crypto} suggests that research in biometric fusion has primarily focused on combining multiple sources of information for different problems and designing new fusion algorithms. The questions of \textit{what}, \textit{when}, and \textit{how} to fuse are important for the development of a biometric fusion system, and need to be answered during algorithm design. However, in order to develop efficient real world biometric fusion systems, they also require efficient implementation and domain adaptation to account for changes in sensor technology, environment, target population, etc. In order to be practically deployable, we believe the following to be some important research topics that require more attention and focused research efforts. \vspace{3pt}


\noindent\textbf{(i) Portability of Multibiometric Solutions:} 
Most fusion algorithms have a number of tunable parameters. For example, even the simple sum rule for score-level fusion requires the estimation of score normalization parameters and the weight vector. Automatically deducing these parameters for different applications is not an easy task. Even learning-based methods for automatically deducing fusion parameters are vulnerable to biases in the training data. Thus, directly transferring the fusion module from one application to another may not be viable in practical systems. The problem is further exacerbated due to differences in sensors, environment, population, etc. across applications. This raises the issue of portability. How can one design robust fusion systems that can be easily ported across applications? 

Domain adaptation and transfer learning have been touted to be effective paradigms for adapting machine learning methods to new application domains ~\cite{patel15Spl,pan10survey,bhatt14Tip}. Research at the intersection of biometric fusion and domain adaptation can have two potential directions: (a) utilizing biometric fusion for domain adaptation, and (b) incorporating domain adaptation in existing multibiometric systems for cross-domain matching. Both these directions have real world applicability with widespread impact in terms of (a) addressing the long standing problem of cross-domain matching, or (b) updating existing multibiometric systems to handle data emerging from fundamentally different distributions. 

\noindent\textbf{(ii) Designing Adaptive and Dynamic Fusion Systems:} 
In real world applications, multibiometric systems often have to operate on large-scale data captured using multiple sensors across different geographical regions from a diverse heterogeneous population (e.g., national ID card program in India). Further, the requirements of an application and the nature of its data may change over time. In the literature, techniques such as online learning or co-training have shown to improve the performance of unibiometric recognition systems by updating them \textit{on the fly} \cite{bhatt14Tip,singh10Image}. However, online learning based techniques are yet to be explored for multibiometric systems. This is an open area of research, i.e., designing fusion methods that continually evolve over time to accommodate changes in system requirements as well as variations in data distribution.  

A pertinent problem is the issue of template update, i.e., modifying the stored biometric data of an individual in order to account for intra-class variations~\cite{UludagTemplateSelection2004,rattani09}. Aging and physical ailment can modify the biometric trait of an individual thereby requiring the enrolled data to be periodically updated. Updating the multibiometric templates of a subject over time can be an arduous task and may inadvertently result in identity creep where an impostor can exploit the template update mechanism to take over the identity of an enrolled subject. An adaptive fusion system should be able to discourage such attacks while still accounting for inevitable changes in data distribution that occur over time.

\noindent\textbf{(iii) Multibiometric Security and Privacy:} 
Research in soft biometrics has established the possibility of deducing additional information about an individual (e.g., age, gender, ethnicity, health condition, genetic disorders, etc.) from their biometric data or template~\cite{dantcheva16Tifs}. While this information can be used to improve recognition accuracy, it can also be deemed to violate the privacy of the subject and, in some cases, can be used for profiling an individual. The availability of multibiometric data corresponding to multiple biometric traits will only increase concerns about compromising the privacy of subjects. It is, therefore, necessary to impart security and privacy to the stored templates. In addition, there must be legislative guarantees that prevent the data from being used beyond the purposes for which it was intended at the time of enrollment.

Recent work in differential privacy in the context of a single biometric modality \cite{chhabra18Ijcai,mirjalili18} could potentially be extended to multibiometric templates. However, {\em guaranteeing} privacy may not be an easy task especially due to the advent of powerful deep learning techniques that can be leveraged for gleaning ancillary information about a subject that was previously not thought to be possible~\cite{Martinez2018}. Another related challenge has to do with the retention of recognition accuracy whist imparting security and privacy. In many cases, the use of privacy preserving or security enhancing schemes results in a degradation in recognition accuracy. Balancing security and privacy with matching accuracy is, therefore, an important challenge that needs to be judiciously resolved. Recent work in homomorphic encryption could potentially be appropriated for this purpose~\cite{GomezHomomorphic2017}. The use of non-biometric cues may also be needed to facilitate privacy and enhance security~\cite{ElliotPrivacy2007}.

The principle of ``signal mixing'' is also being used to impart security and privacy to biometric data. Othman and Ross~\cite{OthmanRossMixing2012,RossOthmanMixing2011} describe a mixing scheme where an input fingerprint image is mixed with another fingerprint (e.g., from a different finger) in order to produce a new mixed image, that obscures the identity of the original fingerprint. This can be viewed as a data-level fusion approach. The researchers also developed a method to fuse two distinct modalities, viz., fingerprint and iris, at the image level~\cite{OthmanRossFingerprintIris2015}. 

\noindent\textbf{(iv) Resolving Conflicts Between Information Sources:} 
The availability of multiple biometric sources and, consequently, multiple pieces of biometric evidence, is not always beneficial. In some cases, the individual biometric sources can offer conflicting decisions  about the identity of a subject. For example, in a bimodal identification system, the face and fingerprint modalities may generate a completely different list of ranked identities; or, in a multimodal verification system, half of the component classifiers might confirm the claimed identity, while the other half might refute the claimed identity. In such scenarios, it is necessary to have a principled way to generate a decision. To address this, it may be necessary to re-acquire the biometric traits of an individual and recompute the decisions. Another possibility is to consider only the outputs of the most reliable sources. However, the reliability of a source (e.g., a matcher) will depend upon multiple factors including the quality of the data and the baseline performance of the matcher itself. Pragmatic methods are needed to handle such operationally relevant situations. Another problem closely related to this is the uncertainty associated with the decisions rendered by individual matchers in a multibiometric framework. Incorporating these uncertainty (or confidence) values in the decision architecture would be essential.

\noindent\textbf{(v) Predicting Scalability of Multibiometric Systems:} 
A number of models have been developed to predict the scalability of a biometric system relying on a single modality~\cite{SchmidLargeDeviations2004}. Such prediction models are needed to evaluate the suitability of a given biometric system for an anticipated large-scale application. Developing such prediction models for multibiometric systems is not easy since these systems rely on multiple sources of information and, therefore, the performance of each source has to be first modeled and then combined with the models associated with the other sources~\cite{WangBhanu2006,NandakumarCorrelation2009}. Alternately, the entire multibiometric system can be characterized using a single model. In either case, the degrees of freedom to be considered can be intractable. Thus, predicting the scalability of a multibiometric system requires much more research and effective models are needed to characterize the complex relationship between individual sources.  

\noindent\textbf{(vi) Sensor Configuration in Multimodal Systems:} 
One of the understudied problems in multimodal biometrics is the placement of sensors in the data acquisition module to maximize recognition performance while minimizing user inconvenience. Consider a multibiometric kiosk that identifies individuals based on their gait, face and fingerprint modalities. As the subject approaches the kiosk, the system uses the gait information from a distance to retrieve a list of potential identities. When the subject is reasonably close to the kiosk, the face image is used to further narrow down the list of matching identities and possibly even determining the exact identity of the subject. The fingerprint sensor is only invoked if the gait and face modalities are unable to uniquely identify the subject. Such a system can arguably improve the throughput of the biometric system. However, camera placement and configuration would be a critical issue in this application. More research is needed to model user behavior in such applications and suitably adjust the placement and position of sensors to ensure that the correct identity can be rapidly determined with limited inconvenience to the user.   

\begin{figure}
\begin{center}
\includegraphics[width=3.2in]{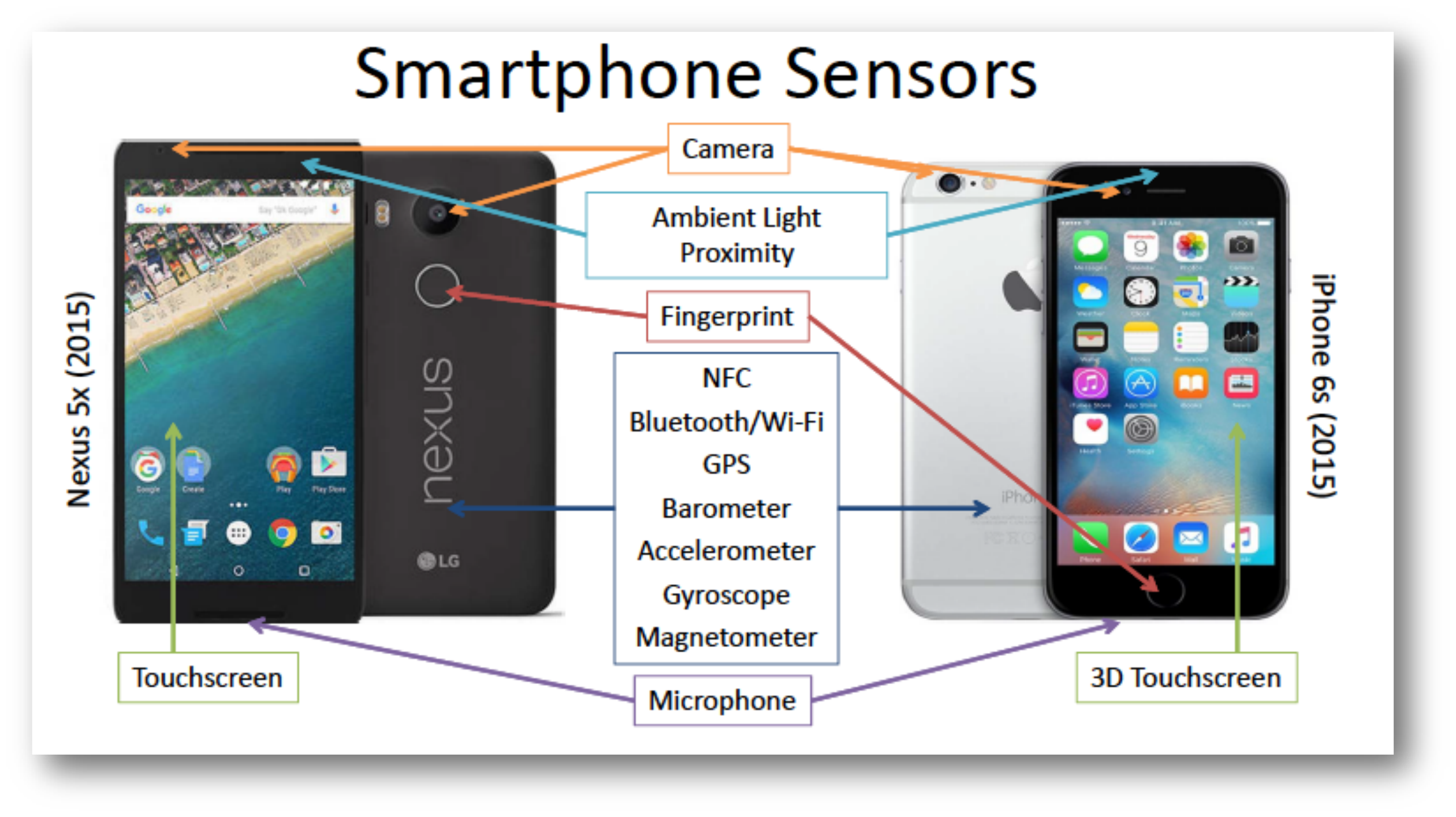}
\caption{Smartphones are equipped with a number of sensors. Data from these sensors can be combined in a judicious manner to perform multimodal user authentication. However, a number of research issues have to be resolved when designing such a solution. \textcopyright Debayan Deb}
\label{fig:smartphone}
\end{center}
\end{figure}

\noindent\textbf{(vii) Multimodal Solutions for Compact Personal Devices:} 
With the increasing use of mobile smartphones and wearable devices, and the need for reliably establishing identity in such compact systems, there is a tremendous opportunity to develop novel biometric sensors. Further, these personal devices are already equipped with a large number of sensors (GPS, accelerometer, gyroscope, magnetometer, microphone, NFC, and heart rate monitors) whose data can be used to identify a subject or to verify an identity in a continuous authentication scheme (Figure \ref{fig:smartphone}). However, principled methods are needed to parse through this heterogeneous data and distill a compact representation that can be used for personal authentication. As described earlier, a number of continuous authentication methods have been developed for personal devices. But these methods typically pre-define the sensor modalities to be used for authentication purposes. Further, they are vulnerable to changes in a subject's behavior and cannot be easily ported from one device to another. Thus, there is a need to develop robust schemes for extracting a distinct and generalizable ``signature" of a subject from the massive amounts of diverse data being generated by devices such as smartphones. 

In summary, while tremendous advances have been made in the field of biometric fusion, it is now time to translate these advancements into operational systems. This provides an unprecedented opportunity for researchers to develop multibiometric solutions that are (a) practically feasible; (b) user friendly; (c) ergonomically tenable; (d) amenable to increased subject throughput; (e) scalable to large heterogeneous populations; (f) compliant with security and privacy requirements; and (g) robust to changes in environment, population, sensors, etc. Such solutions can impact a number of application domains including consumer electronics, banking, autonomous vehicles, robotics, health and medicine, e-commerce, law enforcement, welfare disbursement, border security, national ID cards, cybersecurity and e-voting.

\section*{Acknowledgment}
M. Singh and R. Singh are partially supported through the Infosys CAI at IIIT-Delhi. Ross was supported by the US National Science Foundation under Grant Numbers 1618518 and 1617466 during the writing of this article. 

\vspace{3pt}

\vspace{3pt}

\vspace{3pt}


\section*{References}

\bibliography{mybibfile}

\end{document}